%% file: IVCV20.tex
\RequirePackage{fix-cm}
\documentclass[twocolumn]{svjour3}          
\smartqed  
\usepackage{graphicx}
\usepackage{mathptmx}      

\usepackage{titlesec}

\titleformat{\section}
    {\normalfont\large\bfseries}
    {\thesection}
    {1em}
    {}

\titleformat{\subsection}
  {\bfseries}
  {\thesubsection}
  {1em}
  {}

\titleformat{\subsubsection}
  {\bfseries\itshape}
  {\thesubsubsection}
  {1em}
  {}
  
 \titleformat{\paragraph}[runin]{\itshape}{\theparagraph}{1em}{}
 \titlespacing*{\paragraph}{0pt}{3.25ex plus 1ex minus .2ex}{\the\fontdimen2\font}

\usepackage[nolist]{acronym} 
\input{acronyms} 
\usepackage{amsmath,amssymb} 

\usepackage{booktabs} 
\def\B{\fontseries{b}\selectfont} 
\usepackage{multirow}

\usepackage[dvipsnames]{xcolor}
\newcommand{\etal}{\textit{et al. }}

\usepackage[colorlinks,citecolor=blue,linkcolor=blue]{hyperref}
\usepackage[style=authoryear,natbib=true,maxcitenames=1,uniquelist=false]{biblatex}
\DeclareCiteCommand{\citeyearpar}
    {}
    {\mkbibparens{\bibhyperref{\printfield{year}}\bibhyperref{\printfield{extrayear}}}}
    {\multicitedelim}
    {}
    
\usepackage{dblfloatfix} 

\begin{document} \sloppy

\title{Continuous 3D Multi-Channel Sign Language Production via Progressive Transformers and Mixture Density Networks}

\titlerunning{Continuous 3D Multi-Channel SLP}        

\author{Ben Saunders \and Necati Cihan Camgoz \and Richard Bowden}


\institute{Ben Saunders \and
           Necati Cihan Camgoz \and
           Richard Bowden \at
              Centre for Vision, Speech and Signal Processing, Guildford, UK
}

\date{}


        


\twocolumn[
    \maketitle
    \begin{abstract}
    
        Sign languages are multi-channel visual languages, where signers use a continuous 3D space to communicate. \ac{slp}, the automatic translation from spoken to sign languages, must embody both the continuous articulation and full morphology of sign to be truly understandable by the Deaf community. Previous deep learning-based \ac{slp} works have produced only a concatenation of isolated signs focusing primarily on the manual features, leading to a robotic and non-expressive production.
        
        In this work, we propose a novel Progressive Transformer architecture, the first \ac{slp} model to translate from spoken language sentences to continuous 3D multi-channel sign pose sequences in an end-to-end manner. Our transformer network architecture introduces a counter decoding that enables variable length continuous sequence generation by tracking the production progress over time and predicting the end of sequence. We present extensive data augmentation techniques to reduce prediction drift, alongside an adversarial training regime and a \ac{mdn} formulation to produce realistic and expressive sign pose sequences.
        
        We propose a back translation evaluation mechanism for \ac{slp}, presenting benchmark quantitative results on the challenging PHOENIX14T dataset and setting baselines for future research. We further provide a user evaluation of our \ac{slp} model, to understand the Deaf reception of our sign pose productions.
    
    \end{abstract}
]

\begin{figure*}[t!]
    \centering
    \includegraphics[width=1.0\linewidth]{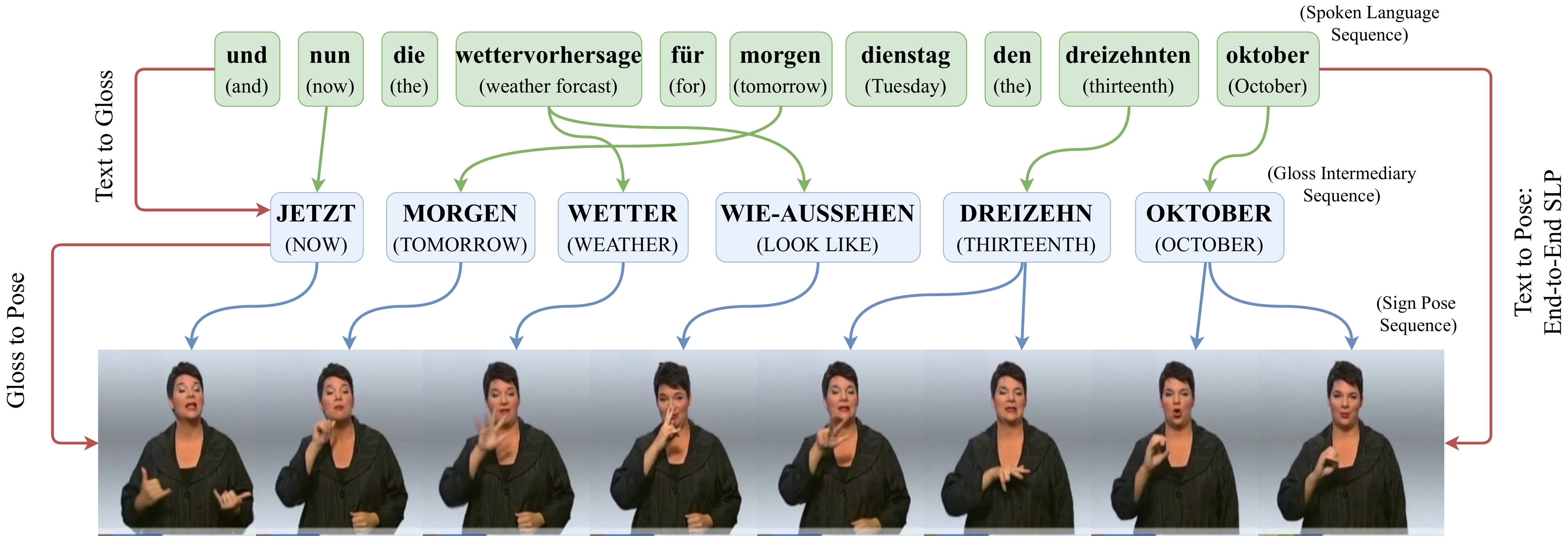}
    \caption{\acf{slp} example showing corresponding spoken language, gloss representation and sign language sequences. The \textit{Text to Gloss}, \textit{Gloss to Pose} and \textit{Text to Pose} translation tasks are highlighted, where end-to-end \ac{slp} is a direct translation from spoken language to sign language, skipping the gloss intermediary. \textit{Note: In this manuscript we use text to denote spoken language sequences.}}
    \label{fig:SLP_Text_Gloss_Seq}
\end{figure*}

\section{Introduction}

Sign languages are visual multi-channel languages and the main medium of communication for the Deaf. Around 5\% of the worlds population experience some form of hearing loss \citep{whoDeaf2020}. In the UK alone, there are an estimated 9 million people who are Deaf or hard of hearing \citep{BDADeaf2020}. For the Deaf native signer, a spoken language may be a second language, meaning their spoken language skills can vary immensely \citep{holt1993stanford}. Therefore, sign languages are the preferred form of communication for the Deaf communities.

Sign languages possess different grammatical structure and syntax to spoken languages \citep{stokoe1980sign}. As highlighted in Figure \ref{fig:SLP_Text_Gloss_Seq}, the translation between spoken and sign languages requires a change in order and structure due to their non-monotonic relationship. Sign languages are also 3D visual languages, with position and movement relative to the body playing an important part of communication. In order to convey complex meanings and context, sign languages employ multiple modes of articulation. The manual features of hand shape and motion are combined with the non-manual features of facial expressions, mouthings and upper body posture \citep{sutton1999linguistics}. 

Sign languages have long been researched by the vision community \citep{tamura1988recognition,starner1997real,bauer2000video}. Previous research has focused on the recognition of sign languages and the subsequent translation to spoken language. Although useful, this is a technology more applicable to allowing the hearing to understand the Deaf, and often not that helpful for the Deaf community. The opposite task of \acf{slp} is far more relevant to the Deaf. Automatically translating spoken language into sign language could increase the sign language content available in the predominately hearing-focused world.


To be useful to the Deaf community, \ac{slp} must produce sequences of natural, understandable sign akin to a human translator \citep{bragg2019sign}. Previous deep learning-based \ac{slp} work has been limited to the production of concatenated isolated signs \citep{stoll2020text2sign,zelinka2020neural}, with a focus solely on the manual features. These works also approach the problem in a fragmented Text to Gloss\footnote{Glosses are a written representation of sign, defined as minimal lexical items.} and Gloss to Pose production (Figure \ref{fig:SLP_Text_Gloss_Seq} left), where important context can be lost in the gloss bottleneck. However, the production of full sign sequences is a more challenging task, as there is no direct alignment between sign sequences and spoken language sentences. Ignoring non-manual features disregards the contextual and grammatical information required to fully understand the meaning of the produced signs \citep{valli2000linguistics}. These works also produce only 2D skeleton data, lacking the depth channel to truly model realistic motion.

In this work, we present a Continuous 3D Multi-Channel Sign Language Production model, the first \ac{slp} network to translate from spoken language sentences to continuous 3D multi-channel sign language sequences in an end-to-end manner. This is shown on the right of Figure \ref{fig:SLP_Text_Gloss_Seq} as a direct translation from source spoken language, without the need for a gloss intermediary. We propose a \textit{Progressive Transformer} architecture that uses an alternative formulation of transformer decoding for continuous sequences, where there is no pre-defined vocabulary. We introduce a counter decoding technique to predict continuous sequences of variable length by tracking the production progress over time and predicting the end of sequence. Our sign pose productions contain both manual and non-manual features, increasing both the realism and comprehension.

To reduce the prediction drift often seen in continuous sequence production, we present several data augmentation methods. These create a more robust model and reduce the erroneous nature of auto-regressive prediction. Continuous prediction often results in a under-articulated output due to the problem of regression to the mean, and thus we propose the addition of adversarial training. A discriminator model conditioned on source spoken language is introduced to prompt a more realistic and expressive sign production from the progressive transformer. Additionally, due to the multimodal nature of sign languages, we also experiment with a \acf{mdn} modelling, utilising the progressive transformer outputs to paramatise a Gaussian mixture model.

To evaluate quantitative performance, we propose a back translation evaluation method for \ac{slp}, using a \ac{slt} back-end to translate sign productions back to spoken language. We evaluate on the challenging \acf{ph14t} dataset, presenting several benchmark results of both \textit{Gloss to Pose} and \textit{Text to Pose} configurations, to underpin future research. We also provide a user evaluation of our sign productions, to evaluate the comprehension of our \ac{slp} model. Finally, we share qualitative results to give the reader further insight into the models performance, producing accurate sign pose sequences of unseen text input.

The contributions of this paper can be summarised as:
\begin{itemize}
    \item The first \ac{slp} model to translate from spoken language to continuous 3D sign pose sequences, enabled by a novel transformer decoding technique
    \item An application of conditional adversarial training to \ac{slp}, for the production of realistic sign
    \item The combination of transformers and mixture density networks to model multimodal continuous sequences
    \item Benchmark \ac{slp} results on the \ac{ph14t} dataset and a new back translation evaluation metric, alongside a comprehensive Deaf user evaluation
\end{itemize}

Preliminary versions of this work were presented in Saunders \etal \citeyearpar{saunders2020adversarial,saunders2020progressive}. This extended manuscript includes additional formulation and the introduction of a \ac{mdn} modelling for expressive sign production. Extensive new quantitative and qualitative evaluation is provided to explore the capabilities of our approach, alongside a user study with Deaf participants to measure the comprehension of our produced sign language sequences.

The rest of this paper is organised as follows: We outline the previous work in \ac{slp} and surrounding areas in Section~\ref{sec:related_work}. Our progressive transformer network and proposed model configurations are presented in Section~\ref{sec:methodology}. Section \ref{sec:setup} provides the experimental setup, with quantitative evaluation in Section~\ref{sec:quant} and qualitative evaluation in Section~\ref{sec:qual}. Finally, we conclude the paper in Section \ref{sec:conc} by discussing our findings and future work.

\section{Related Work} \label{sec:related_work}

To understand the sign language computational research landscape, we first outline the recent literature in \ac{slr} and \ac{slt} and then detail previous work in \ac{slp}. Sign languages reside at the intersection between vision and language, so we also review recent developments in \ac{nmt}. Finally, we provide background on the applications of Adversarial Training and \acfp{mdn} to sequence tasks, specifically applied to human pose generation.

\subsection{Sign Language Recognition \& Translation}
The goal of vision-based sign language research is to develop systems capable of recognition, translation and production of sign languages \citep{bragg2019sign}. There has been prominent sign language computational research for over 30 years \citep{tamura1988recognition,starner1997real,bauer2000video}, with an initial focus on isolated sign recognition \citep{grobel1997isolated,ozdemir2016isolated} and a recent expansion to \ac{cslr} \citep{chai2013sign,koller2015continuous,camgoz2017subunets,}. However, the majority of work has relied on manual feature representations \citep{cooper2012sign} and statistical temporal modelling \citep{vogler1999parallel}. 

Recently, larger sign language datasets have been released, such as \ac{ph14} \citep{forster2014extensions}, Greek Sign Language (GSL) \citep{adaloglou2019comprehensive} and the Chinese Sign Language Recognition Dataset \citep{huang2018video}. These have enabled the application of deep learning approaches to \ac{cslr}, such as \acp{cnn} \citep{koller2016deepsign,koller2019weakly} and \acp{rnn} \citep{cui2017recurrent,koller2017resign}.

Expanding upon \ac{cslr}, Camgoz \etal \citeyearpar{camgoz2018neural} introduced the task of \ac{slt}, aiming to directly translate sign videos to spoken language sentences. Due to the differing grammar and ordering between sign and spoken language \citep{stokoe1980sign}, \ac{slt} is a more challenging task than \ac{cslr}. The majority of work has utilised \ac{nmt} networks for \ac{slt} \citep{camgoz2018neural,ko2019neural,orbay2020neural,yin2020sign}, translating directly to spoken language or via a gloss intermediary. Transformer based models are the current state-of-the-art in \ac{slt}, jointly learning the recognition and translation tasks \citep{camgoz2020sign}. The inclusion of multi-channel features have also been shown to reduce the dependence on gloss annotation in \ac{slt} \citep{camgoz2020multi}.

\subsection{Sign Language Production}
Previous research into \ac{slp} has focused on avatar-based techniques that generate realistic-looking sign production, but rely on pre-recorded phrases that are expensive to create \citep{zwitserlood2004synthetic,glauert2006vanessa,ebling2015bridging,mcdonald2016automated}. Non-manual feature production has been included in avatar generation, such as mouthings \citep{elliott2008linguistic} and head positions \citep{cox2002tessa}, but have been viewed as ``stiff and emotionless'' with an ``absense of mouth patterns'' \citep{kipp2011assessing}. MoCap approaches have successfully produced realistic productions, but are expensive to scale \citep{lu2010collecting}. \ac{smt} has also been applied to \ac{slp} \citep{kouremenos2018statistical,kayahan2019hybrid}, relying on rules-based processing that can be difficult to encode. 

Recently, there has been an increase in deep learning approaches to automatic \ac{slp} \citep{stoll2020text2sign,xiao2020skeleton,zelinka2020neural}. Stoll \etal \citeyearpar{stoll2020text2sign} presented a \ac{slp} model that used a combination of \ac{nmt} and \acp{gan}. The authors break the problem into three independent processes trained separately, producing a concatenation of isolated 2D skeleton poses mapped from sign glosses via a look-up table. As seen with other works, this production of isolated signs of a set length and order without realistic transitions results in robotic animations that are poorly received by the Deaf \citep{bragg2019sign}. Contrary to Stoll \etal, our work focuses on automatic sign production and learning the mapping between text and skeleton pose sequences directly, instead of providing this a priori.

The closest work to this paper is that of Zelinka \etal \citeyearpar{zelinka2020neural}, who use a neural translator to synthesise skeletal pose from text. A single 7-frame sign is produced for each input word, generating sequences with a fixed length and ordering that disregards the natural syntax of sign language. In contrast, our model allows a dynamic length of output sign sequence, learning the length and ordering of corresponding signs from the data, whilst using a progress counter to determine the end of sequence generation. Unlike Zelinka \etal, who work on a proprietary dataset, we produce results on the publicly available \ac{ph14t}, providing a benchmark for future \ac{slp} research.

Previous deep learning-based \ac{slp} works produce solely manual features, ignoring the important non-manuals that convey crucial context and meaning. Mouthings, in particular, are vital to the comprehension of most sign languages, differentiating signs that may otherwise be homophones. The expansion to non-manuals is challenging due to the required temporal coherence with manual features and the intricacies of facial movements. We expand production to non-manual features by generating synchronised mouthings and facial movements from a single model, for expressive and natural sign production.

\subsection{Neural Machine Translation:}

\ac{nmt} is the automatic translation from a source sequence to a target sequence of a differing language, using neural networks. To tackle this sequence-to-sequence task, \acp{rnn} were introduced by Cho \etal \citeyearpar{cho2014learning}, which iteratively apply a hidden state computation across each token of the sequence. This was later developed into encoder-decoder architectures \citep{sutskever2014sequence}, which map both sequences to an intermediate embedding space. Encoder model have the drawback of a fixed sized representation of the source sequence. This problem was overcome by an attention mechanism that facilitated a soft-search over the source sentence for the most useful context \citep{bahdanau2015neural}.

Transformer networks were recently proposed by Vaswani \etal \citeyearpar{vaswani2017attention}, achieving state-of-the-art performance in many \ac{nmt} tasks. Transformers use self-attention mechanisms to generate representations of entire sequences with global dependencies. \ac{mha} layers are used to model different weighted combinations of each sequence, improving the representational power of the model. A mapping between the source and target sequence representations is created by an encoder-decoder attention, learning the sequence-to-sequence task.

Transformers have achieved impressive results in many classic \ac{nlp} tasks such as language modelling \citep{dai2019transformer,zhang2019ernie} and sentence representation \citep{devlin2018bert}, alongside other domains including image captioning \citep{zhou2018end} and action recognition \citep{girdhar2019video}. Related to this work, transformer networks have been applied to many continuous output tasks such as speech synthesis \citep{ren2019fastspeech}, music production \citep{huang2018music} and speech recognition \citep{povey2018time}.

Applying sequence-to-sequence methods to continuous output tasks is a relatively underresearched problem. In order to determine sequence length of continuous outputs, previous works have used a fixed output size \citep{zelinka2020neural}, a binary end-of-sequence (EOS) flag \citep{graves2013generating} or a continuous representation of an EOS token \citep{mukherjee2019predicting}. We propose a novel counter decoding technique that predicts continuous sequences of variable length by tracking the production progress over time and implicitly learning the end of sequence.

\subsection{Adversarial Training}

Adversarial training is the inclusion of a discriminator model designed to improve the realism of a generator by critiquing the productions \citep{goodfellow2014generative}. \acp{gan}, which generate data using adversarial techniques, have produced impressive results when applied to image generation \citep{radford2015unsupervised,isola2017image,zhu2017unpaired} and, more recently, video generation tasks \citep{vondrick2016generating,tulyakov2018mocogan}. Conditional \acp{gan} \citep{mirza2014conditional} extended \acp{gan} with generation conditioned upon specific data inputs.

\acp{gan} have also been applied to natural language tasks \citep{zhang2016generating,lin2017adversarial,press2017language}. Specific to \ac{nmt}, Wu \etal \citeyearpar{wu2017adversarial} designed Adversarial-NMT, complimenting the original \ac{nmt} model with a \ac{cnn} based adversary, and Yang \etal \citeyearpar{yang2017improving} proposed a \ac{gan} setup with translation conditioned on the input sequence.

Specific to human pose generation, adversarial discriminators have been used for the production of realistic pose sequences \citep{cai2018deep,chan2019everybody,ren2019music}. Ginosar \etal \citeyearpar{ginosar2019learning} show that the task of generating skeleton motion suffers from regression to the mean, and adding an adversarial discriminator can improve the realism of gesture production. Lee \etal \citeyearpar{lee2019dancing} use a conditioned discriminator to produce smooth and diverse human dancing motion from music. In this work, we use a conditional discriminator to produce expressive sign pose outputs from source spoken language.

\begin{figure*}[t!]
    \centering
    \includegraphics[width=0.8 \linewidth]{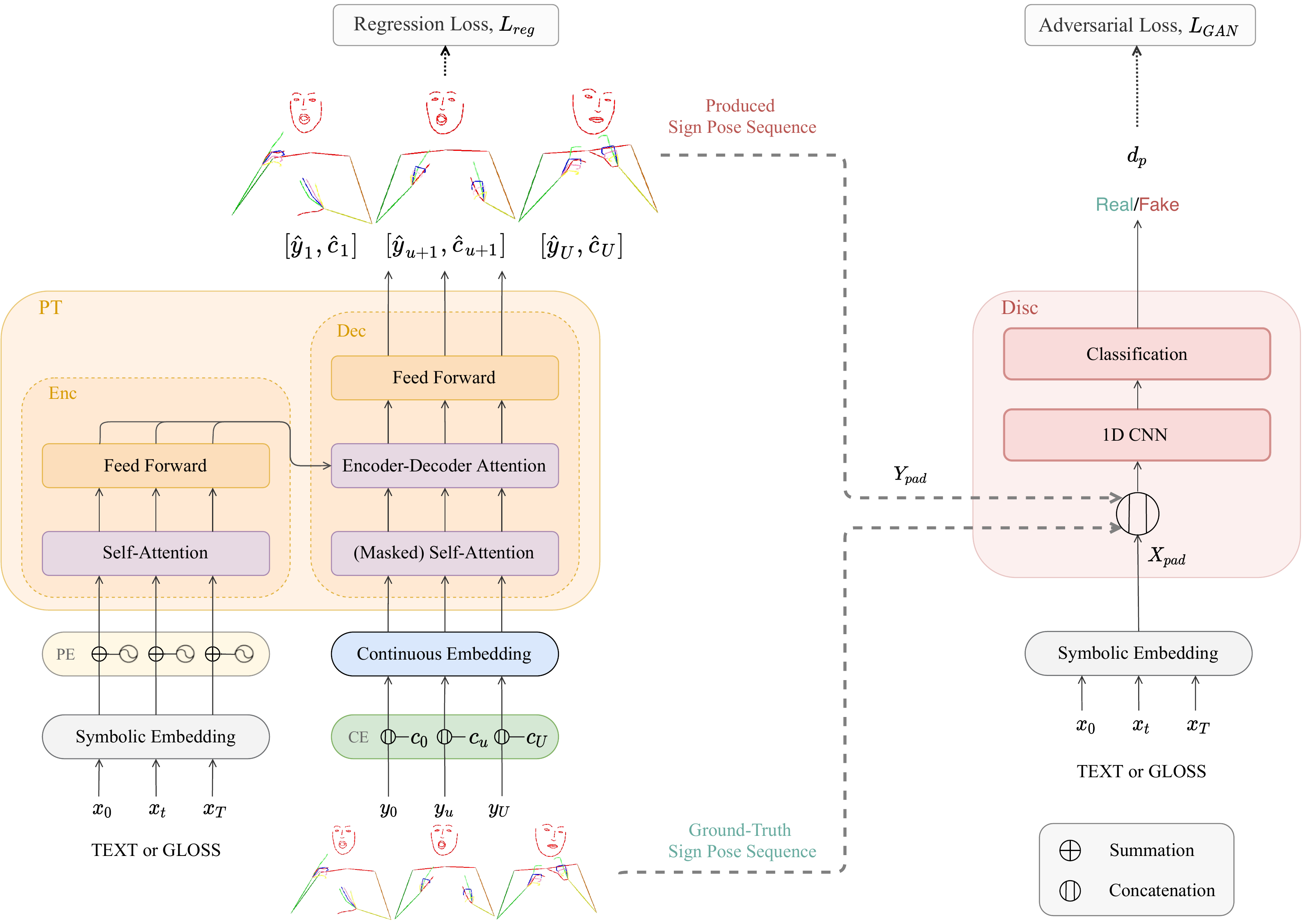}
    \caption{Architecture details of our \textit{Progressive Transformer} and \textit{Conditional Discriminator} network. The \textit{Progressive Transformer} produces a sign pose sequence, $\hat{y}_{1:U}$, and respective counter values, $\hat{c}_{1:U}$, from source spoken language, $\hat{x}_{1:T}$, in an auto-regressive prediction. The \textit{Conditional Discriminator} takes as input either ground-truth or produced sign pose sequences alongside the respective source spoken language, and predicts a single realism scalar, ${d_p}$. The network is trained end-to-end via a weighted combination of regression loss, $L_{reg}$, and adversarial loss, $L_{GAN}$. (\textbf{PT}: Progressive Transformer, \textbf{PE}: Positional Encoding, \textbf{CE}: Counter Encoding, \textbf{Disc}: Discriminator)}
    \label{fig:PT_Model_Overview}
\end{figure*}%
%

\subsection{Mixture Density Networks}

\acfp{mdn} create a multimodal prediction to better model distributions that may not be modelled fully by a single density distribution. \acp{mdn} combine a conventional neural network with a mixture density model, modelling an arbitrary conditional distribution via a direct parametrisation \citep{bishop1994mixture}. The neural network estimates the density components, predicting the weights and statistics of each distribution. 

\acp{mdn} are often used for continuous sequence generation tasks due to their ability to model sequence uncertainty \citep{schuster2000better}. Graves \etal \citeyearpar{graves2013generating} combined an \ac{mdn} with a \ac{rnn} for continuous handwriting generation, which has been expanded to sketch generation \citep{zhang2017drawing,ha2018neural} and reinforcement learning \citep{ha2018recurrent}. \acp{mdn} have also been applied to speech synthesis \citep{wang2017autoregressive}, future prediction \citep{makansi2019overcoming} and driving prediction \citep{hu2018probabilistic}.

\acp{mdn} have also been used for human pose estimation, either to predict multiple hypotheses \citep{li2019generating}, to better model uncertainty \citep{prokudin2018deep,varamesh2020mixture} or to deal with occlusions \citep{ye2018occlusion}. To the best of our knowledge, this work is the first to combine transformers with \acp{mdn} for sequence modelling. We employ \acp{mdn} to capture the natural variability in sign languages and to model production using multiple distributions.

\section{Continuous 3D Sign Language Production}
\label{sec:methodology}

In this section, we introduce our \ac{slp} model, which learns to translate spoken language sentences to continuous sign pose sequences. Our objective is to learn the conditional probability $p(Y|X)$ of producing a sequence of signs \hbox{$Y = (y_{1},...,y_{U})$} with $U$ frames, given a spoken language sentence \hbox{$X = (x_{1},...,x_{T})$} with $T$ words. Glosses could also be used as source input, replacing the spoken language sentence as an intermediary. In this work we represent sign language as a sequence of continuous skeleton poses modelling the 3D coordinates of a signer, of both manual and non-manual features.

Producing a target sign language sequence from a reference spoken language sentence poses several challenges. Firstly, there exists a non-monotic relationship between spoken and sign language, due to the different grammar and syntax in the respective domains \citep{stokoe1980sign}. Secondly, the target signs inhabit a continuous vector space, requiring a differing representation to the discrete space of text and disabling the use of classic end of sequence tokens. Finally, there are multiple channels encompassed within sign that must be produced concurrently, such as the manual (hand shape and position) and non-manual features (mouthings and facial expressions) \citep{pfau2010nonmanuals}.

To address the production of continuous sign sequences, we propose a \textit{Progressive Transformer} model that enables translation from a symbolic to a continuous sequence domain (PT in Figure \ref{fig:PT_Model_Overview}). We introduce a counter decoding that enables the model to track the progress of sequence generation and implicitly learn sequence length given a source sentence. We also propose several data augmentation techniques that reduce the impact of prediction drift.

To enable the production of expressive sign, we introduce an adversarial training regime for \ac{slp}, supplementing the progressive transformer generator with a conditional adversarial discriminator, (Disc in Figure \ref{fig:PT_Model_Overview}). To enhance the capability to model multimodal distributions, we also propose a \ac{mdn} formulation of the \ac{slp} network. In the remainder of this section we describe each component of the proposed architecture in detail.

\begin{figure*}[t!]
    \centering
    \includegraphics[width=0.7 \linewidth]{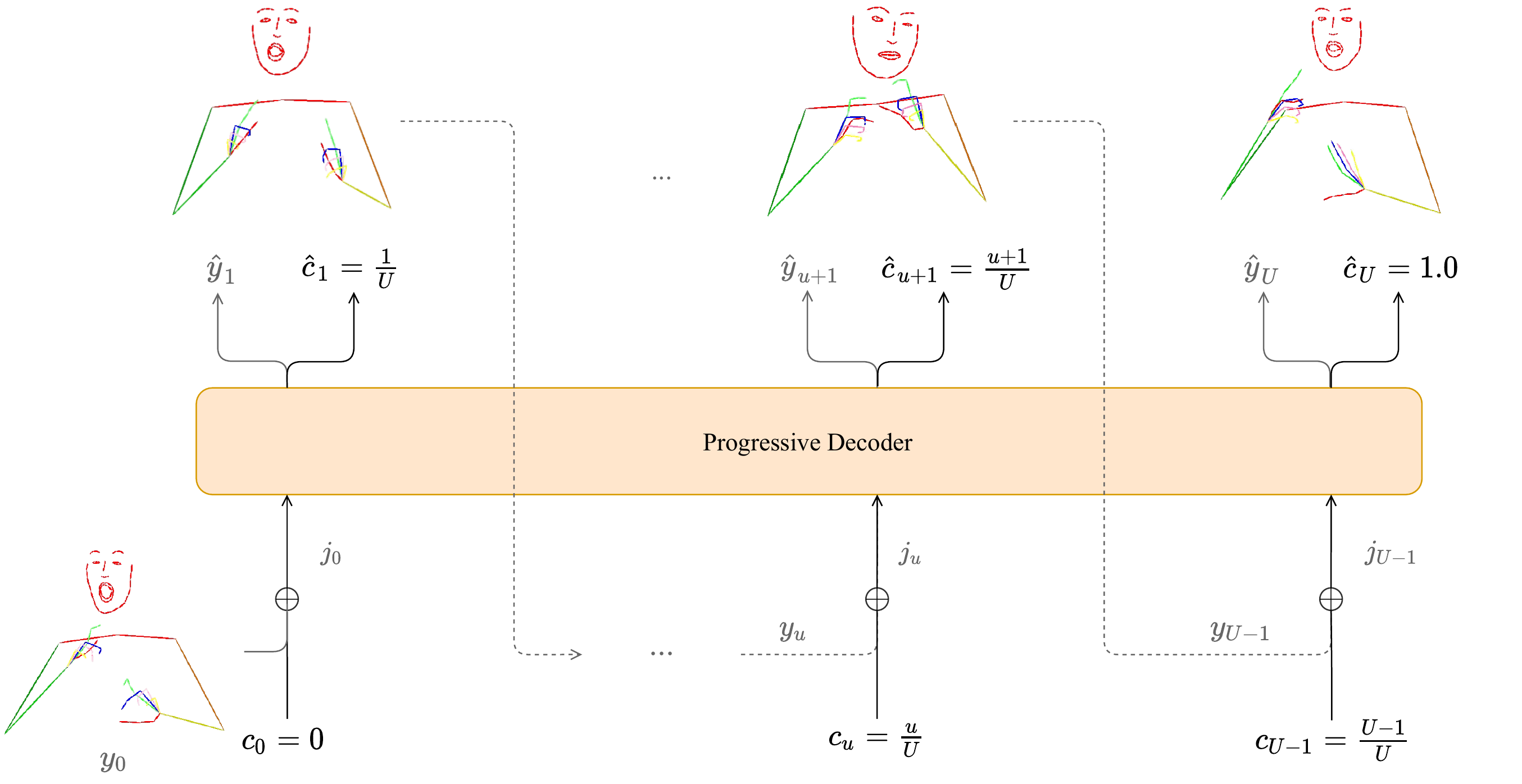}
    \caption{Counter decoding example, showing the simultaneous auto-regressive prediction of continuous sign pose, $\hat{y}_{u}$, and counter value, $\hat{c}_{u} \in \{0:1\}$. A counter value of 1, $\hat{c}=1.0$, denotes end of sequence and decoding is stopped.}
    \label{fig:counter_decoding}
\end{figure*}%

\subsection{Progressive Transformer} \label{sec:progressive}

We build upon the classic transformer \citep{vaswani2017attention}, a model designed to learn the mapping between symbolic source and target languages. We modify the architecture to deal with continuous output representations such as sign language, alongside introducing a counter decoding technique that enables sequence prediction of variable lengths. Our \ac{slp} model tracks the progress of continuous sequence production through time, hence the name \textit{Progressive Transformer}.

In this work, Progressive Transformers translate from the symbolic domains of gloss or spoken language to continuous 3D sign pose sequences. These sequences represent the motion of a signer producing a sign language sentence. The model must produce sign pose outputs that express an accurate translation of the given input sequence and embody a realistic sign pose sequence. Our model consists of an encoder-decoder architecture, where the source sequence is first encoded to a latent representation before being mapped to a target output during decoding in an auto-regressive manner.

\subsubsection{Source Embeddings}

As per the standard \ac{nmt} pipeline, we first embed the symbolic source tokens, $x_{t}$, via a linear embedding layer \citep{mikolov2013distributed}. This represent the one-hot-vector in a higher-dimensional space where tokens with similar meanings are closer. This embedding, with weight, $W$, and bias, $b$, can be formulated as:
\begin{equation}
\label{eq:word_embedding}
    w_{t} = W^{x} \cdot x_{t} + b^{x}
\end{equation}
where $w_{t}$ is the vector representation of the source tokens. 

As with the original transformer implementation, we apply a temporal encoding layer after the source embedding, to provide temporal information to the network. For the encoder, we apply positional encoding, as:
\begin{equation}
\label{eq:word_PE}
    \hat{w}_{t} = w_{t} + \textrm{PositionalEncoding}(t)
\end{equation}
where PositionalEncoding is a predefined sinusoidal function conditioned on the relative sequence position $t$ \citep{vaswani2017attention}. 

\subsubsection{Target Embeddings}

The target sign sequence consists of 3D joint positions of the signer. Due to their continuous nature, we first apply a novel temporal encoding, which we refer to as counter encoding (CE in Figure \ref{fig:PT_Model_Overview}). The counter, $c$, holds a value between 0 and 1, representing the frame position relative to the total sequence length. The target joints, $y_{u}$, are concatenated with the respective counter value, $c_{u}$, formulated as:
\begin{equation}
\label{eq:counter_appending}
    j_{u} = [y_{u},c_{u}]
\end{equation}
where $c_{u}$ is the counter value for frame $u$, as a proportion of sequence length, $U$. At each time-step, counter values, $\hat{c}$, are predicted alongside the skeleton pose, as shown in Figure \ref{fig:counter_decoding}, with sequence generation concluded once the counter reaches 1. We call this process \textit{Counter Decoding}, determining the progress of sequence generation and providing a way to predict the end of sequence without the use of a tokenised vocabulary. 

The counter value provides the model with information relating to the length and speed of each sign pose sequence, determining the sign duration. At inference, we drive the sequence generation by replacing the predicted counter value, $\hat{c}$, with the linear timing information, $c^{*}$, to produce a stable output sequence.

These counter encoded joints, $j_{u}$, are next passed through a linear embedding layer, which can be formulated as:
\begin{equation}
\label{eq:joint_embedding}
    \hat{j}_{u} = W^{y} \cdot j_{u} + b^{y}
\end{equation}
where $\hat{j}_{u}$ is the embedded 3D joint coordinates of each frame, $y_{u}$. 

\subsubsection{Encoder}

The progressive transformer encoder, $E_{PT}$, consists of a stack of $L$ identical layers, each containing 2 sub-layers. Given the temporally encoded source embeddings, $\hat{w}_{t}$, a \ac{mha} sub-layer first generates a weighted contextual representation, performing multiple projections of scaled dot-product attention. This aims to learn the relationship between each token of the sequence and how relevant each time step is in the context of the full sequence. Formally, scaled dot-product attention outputs a vector combination of values, $V$, weighted by the relevant queries, $Q$, keys, $K$, and dimensionality, $d_{k}$:
\begin{equation}
\label{eq:attention}
    \textrm{Attention}(Q,K,V) = \text{softmax}(\frac{Q K^{T}}{\sqrt{d_{k}}})V
\end{equation}

\ac{mha} uses multiple self-attention heads, $h$, to generate parallel mappings of the same queries, keys and values, each with varied learnt parameters. This allows different representations of the input sequence to be generated, learning complementary information in different sub-spaces. The outputs of each head are then concatenated together and projected forward via a final linear layer, as:
\begin{align}
\label{eq:multi_head_attention}
    \textrm{MHA}&(Q,K, V)  =  [head_{1}, ... ,head_{h}] \cdot W^{O}, \nonumber \\
      &  \textrm{where} \medspace head_{i} = \textrm{Attention}(QW_{i}^{Q}, KW_{i}^{K}, VW_{i}^{V})
\end{align}
and $W^{O}$,$W_{i}^{Q}$,$W_{i}^{K}$ and $W_{i}^{V}$ are weights related to each input variable.

The outputs of \ac{mha} are then fed into a second sub-layer of a non-linear feed-forward projection. A residual connection \citep{he2016deep} and subsequent layer norm \citep{ba2016layer} is employed around each of the sub-layers, to aid training. The final encoder output can be formulated as:
\begin{equation}
\label{eq:symbolic_encoder}
    h_{t} = E_{PT}(\hat{w}_{t}  | \hat{w}_{1:T})
\end{equation}
where $h_{t}$ is the contextual representation of the source sequence.

\begin{figure*}[t!]
    \centering
    \includegraphics[width=1.0 \linewidth]{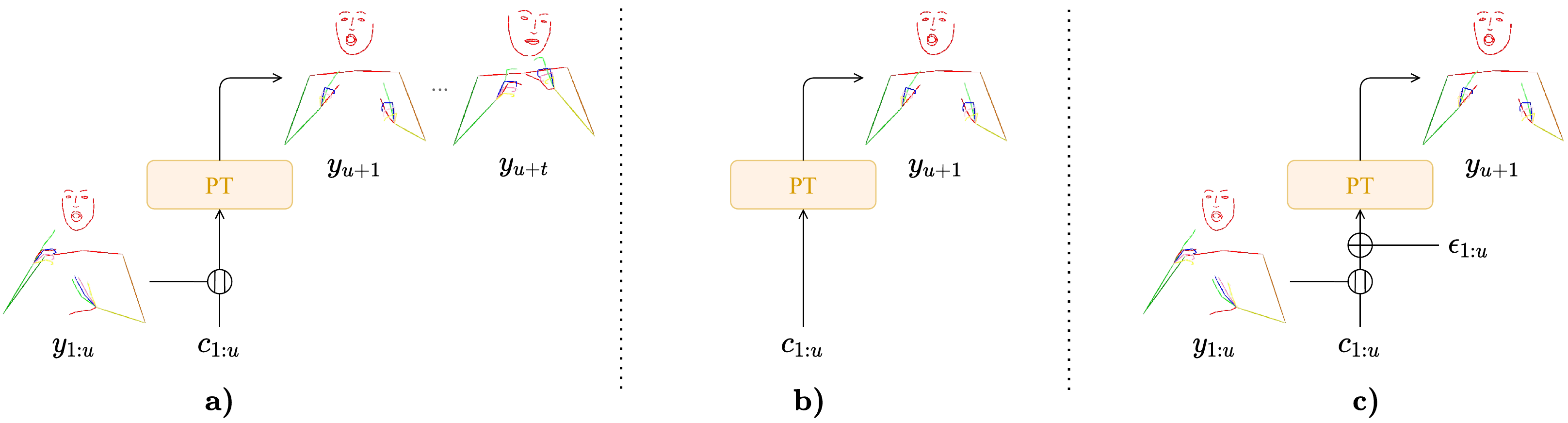}
    \caption{Data augmentation techniques to reduce prediction drift and create a more robust \ac{slp} model. \textbf{a)} Future Prediction is the prediction of multiple future frames. \textbf{b)} Just Counter uses only the counter positions as input. \textbf{c)} Gaussian Noise applies noise to the input skeleton pose. (PT: Progressive Transformer)}
    \label{fig:augmentation}
\end{figure*}%

\subsubsection{Decoder}

The progressive transformer decoder ($D_{PT}$) is an auto-regressive model that produces a sign pose frame at each time-step, alongside the previously described counter value. Distinct from symbolic transformers, our decoder produces continuous sequences. 

The counter-concatenated joint embeddings, $\hat{j}_{u}$, are used to represent the sign pose of each frame. Firstly, an initial \ac{mha} sub-layer is applied to the joint embeddings, similar to the encoder but with an extra masking operation. The masking of future frames prevents the model from attending to subsequent time steps that are yet to be decoded.

A further \ac{mha} mechanism is then used to map the symbolic representations from the encoder to the continuous domain of the decoder. A final feed forward sub-layer follows, with each sub-layer followed by a residual connection and layer normalisation as in the encoder. The output of the progressive decoder can be formulated as:
\begin{equation}
\label{eq:progressive_decoder}
    [\hat{y}_{u},\hat{c}_{u}] = D_{PT}(\hat{j}_{1:u-1} , h_{1:T} )
\end{equation}
where $\hat{y}_{u}$ corresponds to the 3D joint positions representing the produced sign pose of frame $u$ and $\hat{c}_{u}$ is the respective counter value. The decoder learns to generate one frame at a time until the predicted counter value, $\hat{c}_{u}$, reaches 1, determining the end of sequence as seen in Figure \ref{fig:counter_decoding}. The model is trained using the \ac{mse} loss between the predicted sequence, $\hat{y}_{1:U}$, and the ground truth, $y_{1:U}^{*}$:
\begin{equation}
\label{eq:loss_mse}
    L_{MSE} = \frac{1}{U} \sum_{i=1}^{u} ( y_{1:U}^{*} - \hat{y}_{1:U} ) ^{2}
\end{equation}

At inference time, the full sign pose sequence, $\hat{y}_{1:U}$, is produced in an auto-regressive manner, with predicted sign frames used as input to future time steps. Once the predicted counter value reaches 1, decoding is complete and the full sign sequence is produced.

\subsection{Data Augmentation} \label{sec:data_augmentation}

Auto-regressive sequential prediction can often suffer from prediction drift, with erroneous predictions accumulating over time. As transformer models are trained to predict the next time-step using ground truth inputs, they are often not robust to noise in predicted inputs. The impact of drift is heightened for an \ac{slp} model due to the continuous nature of skeleton poses. As neighbouring frames differ little in content, a model can learn to just copy the previous ground truth input and receive a small loss penalty. 

At inference time, with predictions based off previous outputs, errors are quickly propagated throughout the entire sign sequence production. To overcome the problem of prediction drift, in this section we propose various data augmentation approaches, namely \textit{Future Prediction}, \textit{Just Counter} and \textit{Gaussian Noise}.

\subsubsection{Future Prediction} 
Our first data augmentation method is conditional future prediction, requiring the model to predict more than just the next frame in the sequence. Figure \ref{fig:augmentation}a shows an example future prediction of $y_{u+1},...,y_{u+t}$ from the input $y_{1:u}$. Due to the short time step between neighbouring frames, the movement between frames is small and the model can learn to just predict the previous frame with some noise. Predicting more frames into the future means the movement of sign has to be learnt, rather than simply copying the previous frame. At inference time, only the next frame prediction is considered for production.
\subsubsection{Just Counter}
Inspired by the memorisation capabilities of transformer models, we next propose a pure memorisation approach to sign production. Contrary to the usual input of full skeleton joint positions, only the counter values are provided as target input. Figure \ref{fig:augmentation}b demonstrates the input of $c_{1:u}$ as opposed to $y_{1:u}$. The model must decode the target sign pose sequence solely from the counter positions, having no knowledge of the previous frame positions. This halts the reliance on the ground truth joint embeddings it previously had access to, forcing a deeper understanding of the source spoken language and a more robust production. The network setup is also now identical at both training and inference, with the model having to generalise only to new data rather than new prediction inputs.

\begin{figure}[b]
    \centering
    \includegraphics[width=0.5 \linewidth]{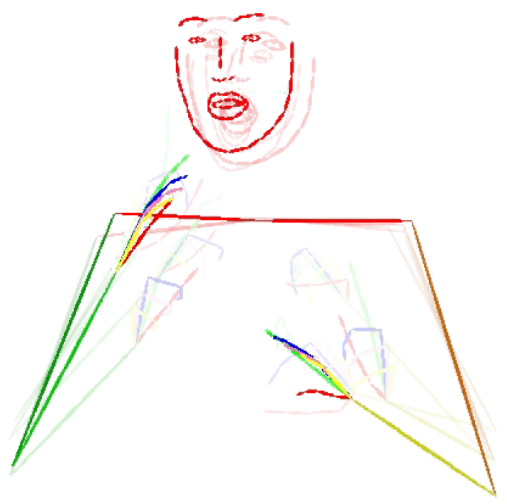}
    \caption{An average of multiple valid sign poses (blurred) results in an under-articulated production due to the problem of regression to the mean.}
    \label{fig:average_skel}
\end{figure}%
\begin{figure*}[t!]
    \centering
    \includegraphics[width=1.0 \linewidth]{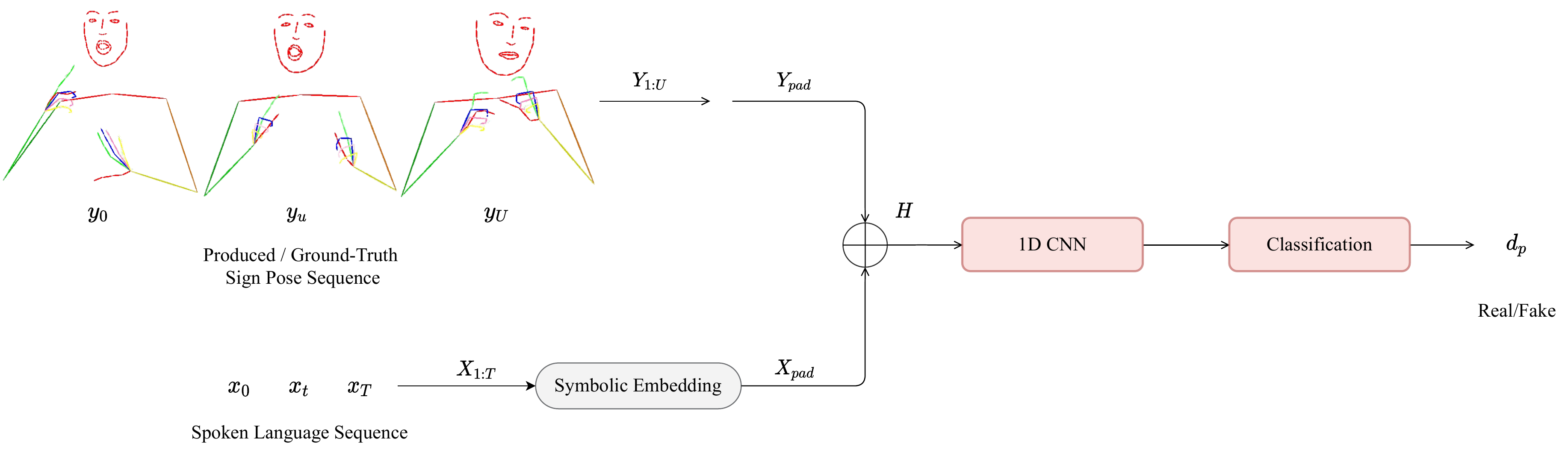}
    \caption{Architecture details of our conditional discriminator model. Sign pose, $Y_{1:U}$, is concatenated with source text, $X_{1:T}$, and projected to a single scalar, $d_{p}$, that represents the realism of the sign pose sequence.}
    \label{fig:discriminator}
\end{figure*}%

\subsubsection{Gaussian Noise}
Our final augmentation technique is the application of noise to the input sign pose sequences during training, increasing the variety of data. This is shown in Figure \ref{fig:augmentation}c, where the input $y_{1:u}$ is summed with noise $\epsilon_{1:u}$. At each epoch, distribution statistics of each joint are collected, with randomly sampled noise applied to the inputs of the next epoch. The addition of Gaussian noise causes the model to become more robust to prediction input error, as it must learn to correct the augmented inputs back to the target outputs. At inference time, the model is more used to noisy inputs, increasing the ability to adapt to erroneous predictions and correct the sequence generation. 

\subsection{Adversarial Training} \label{sec:adversarial}

Sign languages contain naturally varied movements, as each signer produces sign sequences with slightly different articulations and movements. Realistic sign consists of subtle and precise movements of the full body, which can easily be lost when training solely to minimise joint error (e.g. Equation \ref{eq:loss_mse}). \ac{slp} models trained solely for regression can lack pose articulation, suffering from the problem of regression to the mean. Specifically, average hand shapes are produced with a lack of comprehensive motion, due to the high variability of these joints. Figure \ref{fig:average_skel} highlights this problem, as the average of the valid blurred poses results in an under-articulated mean production that does not convey the required meaning.

To address under-articulation, we propose an adversarial training mechanism for \ac{slp}. As shown in Figure \ref{fig:PT_Model_Overview}, we introduce a conditional discriminator, $D$, alongside the \ac{slp} generator, $G$. We frame \ac{slp} as a min-max game between the two networks, with $D$ evaluating the realism of $G$'s productions. We use the previously described progressive transformer architecture as $G$ (Figure \ref{fig:PT_Model_Overview} left) to produce sign pose sequences. We build a convolutional network for $D$ (Figure \ref{fig:discriminator}), trained to produce a single scalar that represents realism, given a sign pose sequence and corresponding source input sequence. These models are co-trained in an adversarial manner, which can be formalised as:

\begin{align}
\label{eq:loss_gan}
    \min_{G}  \max_{D} & \;  L_{GAN}(G,D) = \nonumber \\
      &  \mathbb{E} [\log D(Y^{*} \mid X)] + \mathbb{E} [\log (1-D(G(X) \mid X))] 
\end{align}
where $Y^{*}$ is the ground truth sign pose sequence, $y_{1:U}^{*}$, $G(X)$ equates to the produced sign pose sequence, $\hat{Y} = \hat{y}_{1:U}$, and $X$ is the source spoken language.

\subsubsection{Generator}

Our generator, $G$, learns to produce sign pose sequences given a source spoken language sequence, integrating the progressive transformer into a \ac{gan} framework. Contrary to the standard \ac{gan} implementation, we require sequence generation to be conditioned on a specific source input. Therefore, we remove the traditional noise input \citep{goodfellow2014generative}, and generate a sign pose sequence conditioned on the source sequence, taking inspiration from conditional \acp{gan} \citep{mirza2014conditional}.

We propose training $G$ using a combination of loss functions, namely regression loss, $L_{Reg}$, (Equation \ref{eq:loss_mse}) and adversarial loss, $L^{G}_{GAN}$, (Equation \ref{eq:loss_gan}). The total loss function is a weighted combination of these losses, as:
\begin{equation} \label{eq:loss_total}
    L^{G} = \lambda_{Reg} L_{Reg}(G) + \lambda_{GAN} L^{G}_{GAN}(G,D) 
\end{equation}
where $\lambda_{Reg}$ and $\lambda_{GAN}$ determine the importance of each loss function during training. 

\subsubsection{Discriminator}

\begin{figure*}[h]
    \centering
    \includegraphics[width=0.9 \linewidth]{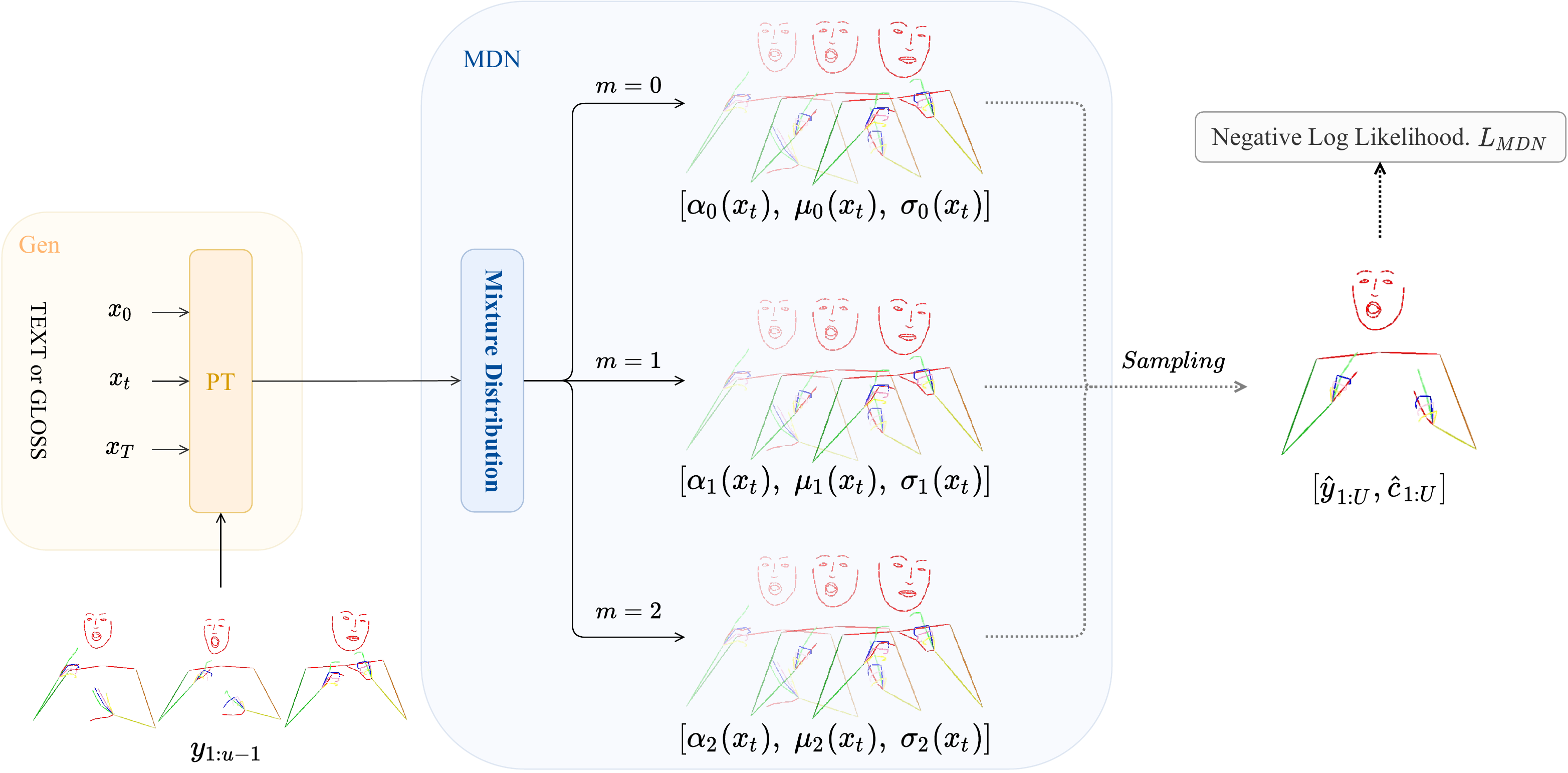}
    \caption{An overview of our \acf{mdn} network. Multiple mixture distributions, $m$, are parameterised by the progressive transformer (PT) outputs, taking input source spoken language and previous sign pose frames. An output sign pose is sampled from the mixture distributions, producing an expressive and variable sign language sequence. The network is trained end-to-end with a negative log likelihood, $L_{MDN}$.}
    \label{fig:mdn}
\end{figure*}%

We present a conditional adversarial discriminator, $D$, used to differentiate generated sign pose sequences, $\hat{Y}$, and ground-truth sign pose sequences, $Y^{*}$, conditioned on the source spoken language sequence, $X$. Figure \ref{fig:discriminator} shows an overview of the discriminator architecture.

For each pair of source-target sequences, $(X,Y)$, of either generated or real sign pose, the aim of $D$ is to produce a single scalar, $d_{p} \in (0,1)$. This represents the probability that the sign pose sequence originates from the data, $Y^{*}$:
\begin{equation} \label{eq:discriminator_scalar}
    d_{p} = P(Y = Y^{*} \mid X,Y) \in (0,1)  
\end{equation}
The sequence counter value is removed before being input to the discriminator, in order to critique only the sign content. Due to the variable frame lengths of the sign sequences, we apply padding to transform them to a fixed length, $U_{max}$, the maximum frame length of target sequences found in the data:
\begin{equation} \label{eq:target_pad}
    Y_{pad} = [Y_{1:U} , \varnothing_{U:U_{max}}]
\end{equation}
where $Y_{pad}$ is the sign pose sequence padded with zero vectors, $\varnothing$, enabling convolutions upon the now fixed size tensor. In order to condition $D$ on the source spoken language, we first embed the source tokens via a linear embedding layer. Again to deal with variable sequence length, these embeddings are also padded to a fixed length $T_{max}$, the maximum source sequence length:
\begin{equation} \label{eq:source_pad}
    X_{pad} = [W^{X} \cdot X_{1:T} + b^{X}, \varnothing_{T:T_{max}}]
\end{equation}
where $W^{X}$ and $b^{X}$ are the weight and bias of the source embedding respectively and $\varnothing$ is zero padding. As shown in the centre of Figure \ref{fig:discriminator}, the source representation is then concatenated with the padded sign pose sequence, to create the conditioned features, $H$:
\begin{equation} \label{eq:discriminator_input}
    H = [Y_{pad}, X_{pad}]
\end{equation}

$N$ 1D convolutional filters are passed over the sign pose sequence, analysing the local context to determine the temporal continuity of the signing motion. This is more effective than a frame level discriminator at determining realism, as a mean hand shape is a valid pose for a single frame, but not consistently over a large temporal window. Leaky ReLU activation \citep{maas2013rectifier} is applied after each layer, promoting healthy gradients during training. A final feed-forward linear layer and sigmoid activation projects the combined features down to the single scalar, $d_{p}$, representing the probability that the sign pose sequence is real.

We train $D$ to maximise the likelihood of producing $d_{p} = 1$ for real sign sequences and $d_{p} = 0$ for generated sequences. This objective can be formalised as maximising Equation \ref{eq:loss_gan}, resulting in the loss function $L^{D} = L^{D}_{GAN}(G,D)$.  At inference time, $D$ is discarded and $G$ is used to produce sign pose sequences in an auto-regressive manner as in Section \ref{sec:progressive}.

\subsection{Mixture Density Networks} \label{sec:mdn}

The previously-described model architectures generate deterministic productions, with each model predicting a single non-stochastic pose at each time step. A single prediction is unable to model any uncertainty or variation that is found in continuous sequence generation tasks like \ac{slp}. The deterministic modelling of sequences can again result in a mean, under-articulated production with no room for expression or variability.


To overcome the issues of deterministic prediction, we propose the use of a \acf{mdn} to model the variation found in sign language. As shown in Figure \ref{fig:mdn}, multiple distributions are used to parameterise the entire prediction subspace, with each mixture component modelling a separate valid movement into the future. This enables prediction of all valid signing motions and their corresponding uncertainty, resulting in a more expressive production.

\subsubsection{Formulation}

\acp{mdn} use a neural network to parameterise a mixture distribution \citep{bishop1994mixture}. A subset of the network predicts the mixture weights whilst the rest generates the parameters of each of the individual mixture distributions. We use our previously described progressive transformer architecture, but amend the output to model a mixture of Gaussian distributions. Given a source token, $x_{t}$, we can model the conditional probability of producing the sign pose frame, $y_{u}$, as:
\begin{equation} \label{eq:MDN_conditional_probability}
    p(y_{u}|x_{t}) = \sum_{i=1}^{M} \alpha_{i}(x_{t}) \phi_{i}(y_{u}|x_{t}) 
\end{equation}
where $M$ is the number of mixture components used in the \ac{mdn}. $\alpha_{i}(x_{t})$ is the mixture weight of the $i^{th}$ distribution, regarded as a prior probability of the sign pose frame being generated from this mixture component. $\phi_{i}(y_{u}|x_{t}) $ is the conditional density of the sign pose for the $i^{th}$ mixture, which can be expressed as a Gaussian distribution:

\begin{equation} \label{eq:MDN_gaussian_dist}
    \phi_{i}(y_{u}|x_{t}) = \frac{1} { \sigma_{i}(x_{t}) \sqrt{2\pi} } 
\; exp^{ \frac{\left \| y_{u} - \mu_{i}(x_{t})  \right \| ^2}{2 \sigma_{i}(x_{t})^2} }
\end{equation}
where $\mu_{i}(x_{t})$ and $\sigma_{i}(x_{t})$ denote the mean and variance of the $i^{th}$ distribution, respectively. The parameters of the \ac{mdn} are predicted directly by the progressive transformer, as shown in Figure \ref{fig:mdn}. The mixture coefficients, $\alpha(x_{t})$, are passed through a softmax activation function to ensure each lies in the range $[0,1]$ and sum to 1. An exponential function is applied to the variances, $\sigma(x_{t})$, to ensure a positive output.

\subsubsection{Optimisation}

During training, we minimise the negative log likelihood of the ground truth data coming from our predicted mixture distribution. This can be formulated as:
\begin{align} \label{eq:MDN_loss}
    L_{MDN} & =  - \sum_{u=1}^{U}  \mathrm{log} \; p(y_{u}|x_{t})  \nonumber \\
     & = - \sum_{u=1}^{U}  \mathrm{log} \sum_{i=1}^{M} \alpha_{i}(x_{t}) \phi_{i}(y_{u}|x_{t}) 
\end{align}
where $U$ is the number of frames in the produced sign pose sequence and $M$ is the number of mixture components. 

\subsubsection{Sampling}

At inference time, we sample sign pose productions from the mixture density computed in Equation \ref{eq:MDN_conditional_probability}, as shown in Figure \ref{fig:mdn}. Firstly, we select the most likely distribution for this source token, $x_{t}$, from the mixture weights, $i_{max} = argmax_{i} \; \alpha_{i}(x_{t})$. From this chosen distribution, we sample the sign pose, predicting $\mu_{i_{max}}(x_{t})$ as a valid pose. To ensure there is no jitter in the sign pose predictions, we set $\sigma(x_{t}) = 0$. This avoids the large variation in small joint positions a large sigma would create, particularly for the hands.

To predict a sequence of multiple time steps, we sample each frame from the mixture density model in an auto-regressive manner as in Section \ref{sec:progressive}. The sampled sign frames are used as input to future transformer time-steps, to produce the full sign pose sequence, $\hat{y}_{1:U}$.

\subsubsection{MDN + Adversarial} \label{sec:mdn_adv}

The \ac{mdn} can also be combined with our adversarial training regime outlined in Section \ref{sec:adversarial}. The \ac{mdn} model is formulated as the adversarial generator pitched against an unchanged conditional discriminator, where a sampled sign pose is used as discriminator input. Again, the final loss function is a weighted combination of the negative log-posterior loss (Equation \ref{eq:MDN_loss}) and the adversarial generator loss (Equation \ref{eq:loss_gan}), as:
\begin{equation} \label{eq:mdn_adv_loss_total}
    L_{MDN}^{G} = \lambda_{MDN} L_{MDN}(G) + \lambda_{GAN} L^{G}_{GAN}(G,D) 
\end{equation}
At inference time, the discriminator model is discarded and a sign pose sequence is sampled from the resulting mixture distribution, as previously explained.

\subsection{Sign Pose Sequence Outputs} \label{sec:sign_seq_outputs}

Each of these model configurations are trained to produce a sign pose sequence, $\hat{y}_{1:U}$, given a source spoken language input, $x_{1:T}$. Animating a video from this skeleton sequence is a trivial task, plotting the joints and connecting the relevant bones, with timing information provided from the progressive transformer counter. These 3D joints can subsequently be used to animate an avatar \citep{kipp2011sign,mcdonald2016automated} or condition a \ac{gan} \citep{chan2019everybody}.

Even though the produced sign pose sequence is a valid translation of the given text, it may be signed at a different speed than that found in the reference data. This is not incorrect, as every signer signs with a varied motion and speed, with our model having its own cadence. However, in order to ease the visual comparison with reference sequences, we apply \ac{dtw} \citep{berndt1994dtw} to temporally align the produced sign pose sequences. This action does not amend the content of the productions, only the temporal coherence for visualisation.

Although our focus has not been on building a real-time system, our current implementation is near real-time and a spoken language sentence can be translated to a sign language video within seconds. However, the nature of translation requires a delay as the context of a whole sentence is needed before it can be translated. As such, the small delay introduced by the automatic system does not present a significant further delay.

\section{Experimental Setup}
\label{sec:setup}

In this section, we outline our experimental setup, detailing the dataset, evaluation metrics and model configuration. We also introduce the back translation evaluation metric and evaluation protocols.

\subsection{Dataset}

In this work, we use the publicly available \ac{ph14t} dataset introduced by Camgoz \etal \citeyearpar{camgoz2018neural}, a continuous \ac{slt} extension of the original \ac{ph14} corpus \citep{forster2014extensions}, becoming the benchmark for \ac{slt} research. This corpus includes parallel \ac{dgs} videos and German translation sequences with redefined segmentation boundaries generated using the forced alignment approach of Koller \etal \citeyearpar{koller2016deepsign}. 8257 videos of 9 different signers are provided, with a vocabulary of 2887 German words and 1066 different sign glosses. We use the original training, validation and testing split as proposed by Camgoz \etal \citeyearpar{camgoz2018neural}.

\begin{figure}[t!]
    \centering
    \includegraphics[width=0.99 \linewidth]{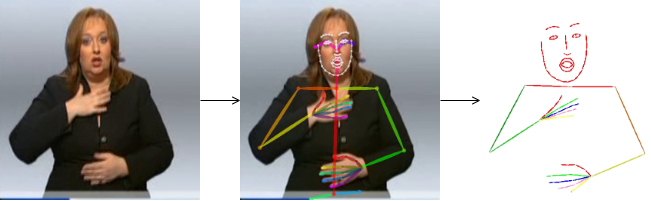}
    \caption{Skeleton pose extraction, using 2D human pose estimation \citep{cao2018openpose} and 2D to 3D mapping \citep{zelinka2020neural}}
    \label{fig:skel_example}
\end{figure}%

We train our \ac{slp} network to generate sequences of 3D skeleton pose representing sign language, as shown in Figure \ref{fig:skel_example}. 2D upper body joint and facial landmark positions are first extracted using OpenPose \citep{cao2018openpose}. We then use the skeletal model estimation improvements presented in Zelinka \etal \citeyearpar{zelinka2020neural} to lift the 2D upper body joint positions to 3D. Finally, we apply skeleton normalisation similar to Stoll \etal \citeyearpar{stoll2020text2sign}, with face coordinates scaled to a consistent size and centered around the nose joint.

\subsection{Back Translation Evaluation}

The evaluation of a continuous sequence generation model is a difficult task, with previous \ac{slp} evaluation metrics of \ac{mse} \citep{zelinka2020neural} falling short of a true measure of sign understanding. In this work, we propose back-translation as a means of \ac{slp} evaluation, translating back from the produced sign pose sequences to spoken language. This provides an automatic measure of how understandable the productions are, and the amount of translation content that is preserved. We find a close correspondence between back translation score and the visual production quality and liken it to the wide use of the inception score for generative models which uses a pre-trained classifier \citep{salimans2016improved}. Similarly, recent \ac{slp} work has used an \ac{slr} discriminator to evaluate isolated skeletons \citep{xiao2020skeleton}, but did not measure the translation performance. Back translation is a relative evaluation metric, best used to compare between similar model configurations. If the chosen \ac{slt} model is amended, absolute model performances will likely also change. However, as we have seen in our experimentation, the relative performance comparisons between models remain consistent. This ensures that comparison results between models remains valid.

We use the state-of-the-art \ac{slt} system \citep{camgoz2020sign} as our back translation model, modified to take sign pose sequences as input. We build a sign language transformer model with 1 layer, 2 heads and an embedding size of 128. This is also trained on the \ac{ph14t} dataset, ensuring a robust translation from sign to text. We generate spoken language translations of the produced sign pose sequences and compute BLEU and ROUGE scores. We provide BLEU n-grams from 1 to 4 for completeness.

\begin{table}[t]
\centering
\caption{Ground-truth back translation results for \textit{Manual}, \textit{Non-Manual} and \textit{Manual + Non-Manual} skeleton pose representations.}
\resizebox{0.8\linewidth}{!}{%
\begin{tabular}{@{}p{0.0cm}c|c@{}}
\toprule
  & \multicolumn{1}{c}{DEV SET} & \multicolumn{1}{c}{TEST SET} \\ 
\multicolumn{1}{r|}{Representation:}      & BLEU-4         & BLEU-4   \\ \midrule
\multicolumn{1}{r|}{Manual} & 11.05 & 9.97 \\
\multicolumn{1}{r|}{Non-Manual} & 8.65 & 9.18 \\
\multicolumn{1}{r|}{Manual + Non-Manual} & {\B 11.44} & {\B 11.01} \\
\bottomrule
\end{tabular}%
}
\label{tab:BT_Val}
\end{table}

We build multiple \ac{slt} models trained with various skeleton pose representations, namely \textit{Manual} (Body), \textit{Non-Manual} (Face) and \textit{Manual + Non-Manual}. We evaluate the back translation performance for each configuration, to see how understandable the representation is and the amount of spoken language that can be recovered. As seen in Table \ref{tab:BT_Val}, the \textit{Manual + Non-Manual} configuration achieves the best back translation result, with \textit{Non-Manual} achieving a significantly lower result. This demonstrates that manual and non-manual features contain complementary information when translating back to spoken language and supports our use of a multi-channel sign pose representation.



\begin{table*}[t!]
\centering
\caption{Text to Gloss translation results of our transformer architecture, compared to that of Stoll \etal \citeyearpar{stoll2020text2sign}.}
\resizebox{0.9\linewidth}{!}{%
\begin{tabular}{@{}p{2.8cm}ccccc|ccccc@{}}
\toprule
            & \multicolumn{5}{c}{DEV SET} & \multicolumn{5}{c}{TEST SET} \\ 
\multicolumn{1}{c|}{Approach:}      & BLEU-4         & BLEU-3         & BLEU-2         & BLEU-1        & ROUGE          & BLEU-4        & BLEU-3         & BLEU-2         & BLEU-1         & ROUGE          \\ \midrule
\multicolumn{1}{r|}{Stoll \etal \citeyearpar{stoll2020text2sign}} & 
16.34       & 22.30         & 32.47        & 50.15       & 48.42    & 15.26      & 21.54       & 32.25        & 50.67        & 48.10          \\
\multicolumn{1}{r|}{Ours}  &  {\B 20.23} & {\B 27.36} & {\B 38.21} & {\B 55.65} & {\B 55.41} & {\B 19.10} & {\B 26.24} & {\B 37.10} & {\B 55.18} & {\B 54.55} \\ \bottomrule
\end{tabular}%
}
\label{tab:text_to_gloss_results}
\end{table*}

As seen in our quantitative experiments in Section \ref{sec:quant}, our sign production sequences can achieve better back translation performance than the original ground truth skeleton data. We believe this is due to a smoothing of the training data during production, as the original data contains artifacts either from 2D pose estimation, the 2D-to-3D mapping or the quality of the data itself. As our model learns to generate a temporally continuous production without these artifacts, our sign pose is significantly smoother than the ground truth. This explains the higher back translation performance from production compared to the ground truth data.

\subsection{Evaluation Protocols} \label{sec:eval_protols}

With back translation as an evaluation metric, we now set \ac{slp} evaluation protocols on the \ac{ph14t} dataset. These can be used as measures for ablation studies and benchmarks for future work. 

\paragraph{\textbf{Text to Gloss (T2G):}}

The first evaluation protocol is the symbolic translation between spoken language and sign language representation. This task is a measure of the translation into sign language grammar, an initial task before a pose production. This can be measured with a direct BLEU and ROUGE comparison, without the need for back translation.

\paragraph{\textbf{Gloss to Pose (G2P)}:}

The second evaluation protocol evaluates the \acp{slp} models capability to produce a continuous sign pose sequence from a symbolic gloss representation. This task is a measure of the production capabilities of a network, without requiring translation from spoken language.

\paragraph{\textbf{Text to Pose (T2P)}:}

The final evaluation protocol is full end-to-end translation from a spoken language input to a sign pose sequence. This is the true measure of the performance of an \ac{slp} system, consisting of jointly performing translation to sign and a production of the sign sequence. Success on this task enables \ac{slp} applications in domains where expensive gloss annotation is not available.

\subsection{Model Configuration}

In the following experiments, our progressive transformer model is built with 2 layers, 4 heads and an embedding size of 512, unless stated otherwise. All parts of our network are trained with Xavier initialisation from scratch \citep{glorot2010understanding}, Adam optimization with default parameters \citep{kingma2014adam} and a learning rate of $10^{-3}$. We use a plateau learning rate scheduler with a patience of 7 epochs, a decay rate of 0.7 and a minimum learning rate of $2\times10^{-4}$. Our code is based on Kreutzer \etal's NMT toolkit, JoeyNMT \citeyearpar{JoeyNMT}, and implemented using PyTorch \citep{paszke2017automatic}. 

\begin{table*}[t!]
\caption{Future Prediction results on the Gloss to Pose task. Evaluation upon modifying the prediction frames, $F_{f}$ to $F_{t}$.}
\centering
\resizebox{0.9\linewidth}{!}{%
\begin{tabular}{@{}p{1.2cm}p{1.2cm}|ccccc|ccccc@{}}
\toprule
     & \multicolumn{5}{c}{DEV SET}  & \multicolumn{5}{c}{TEST SET} \\ 
$F_{f}$ & $F_{t}$ & BLEU-4         & BLEU-3         & BLEU-2         & BLEU-1         & ROUGE          & BLEU-4         & BLEU-3         & BLEU-2         & BLEU-1         & ROUGE          \\ \midrule
0 (Base) & 1 (Base) & 7.38 & 9.62 & 13.81 & 25.03 & 26.55 & 7.13 & 9.30 & 13.63 & 24.86 & 26.03 \\ 
0 & 2 & 9.52 & 12.13 & 16.91 & 27.98 & 30.68 & 9.34 & 11.99 & 16.78 & 28.03 & 30.29 \\ 
0 & 5 & \textbf{11.30} & \textbf{14.17} & 19.19 & 30.45 & \textbf{33.18} & \textbf{10.69} & \textbf{13.49} & \textbf{18.68} & 30.69 & 31.78 \\
0 & 10 & 10.99 & 13.83 & 19.02 & 30.57 & 32.34 & 9.93 & 12.50 & 17.49 & 28.94 & 30.86 \\ 
0 & 20 & 10.08 & 12.84 & 17.79 & 29.30 & 31.27 & 9.23 & 12.02 & 17.27 & 29.53 & 30.11 \\ 
2 & 5 & 10.93 & 13.85 & \textbf{19.23} & \textbf{31.55} & 32.80 & 10.23 & 13.13 & 18.60 & \textbf{30.87} & \textbf{32.38} \\ 
5 & 10 & 10.32 & 13.07 & 18.44 & 30.95 & 31.81 & 9.37 & 12.12 & 17.53 & 30.39 & 30.52 \\ 
\bottomrule
\end{tabular}%
}
\label{tab:future_prediction}
\end{table*}

\begin{table*}[t!]
\caption{Just Counter results on the Gloss to Pose task. Evaluation against a base architecture that uses full skeleton pose as input.}
\centering
\resizebox{0.9\linewidth}{!}{%
\begin{tabular}{@{}p{2.8cm}|ccccc|ccccc@{}}
\toprule
     & \multicolumn{5}{c}{DEV SET}  & \multicolumn{5}{c}{TEST SET} \\ 
\multicolumn{1}{r|}{Configuration:} & BLEU-4         & BLEU-3         & BLEU-2         & BLEU-1         & ROUGE          & BLEU-4         & BLEU-3         & BLEU-2         & BLEU-1         & ROUGE          \\ \midrule
\multicolumn{1}{r|}{Base} & 7.38 & 9.62 & 13.81 & 25.03 & 26.55 & 7.13 & 9.30 & 13.63 & 24.86 & 26.03 \\ 
\multicolumn{1}{r|}{Just Counter} & \textbf{12.34} & \textbf{15.04} & \textbf{21.17} & \textbf{32.43} & \textbf{35.59} & \textbf{12.16} & \textbf{15.50} & \textbf{21.45} & \textbf{33.53} & \textbf{34.80} \\ 

\bottomrule
\end{tabular}%
}
\label{tab:just_counter}
\end{table*}

\begin{table*}[t!]
\caption{Gaussian Noise results on the Gloss to Pose task. Evaluation upon modifying the noise rate, $r_{n}$.}
\centering
\resizebox{0.9\linewidth}{!}{%
\begin{tabular}{@{}p{2.8cm}|ccccc|ccccc@{}}
\toprule
     & \multicolumn{5}{c}{DEV SET}  & \multicolumn{5}{c}{TEST SET} \\ 
 \multicolumn{1}{r|}{$r_{n}$} & BLEU-4         & BLEU-3         & BLEU-2         & BLEU-1         & ROUGE          & BLEU-4         & BLEU-3         & BLEU-2         & BLEU-1         & ROUGE          \\ \midrule
\multicolumn{1}{r|}{0 (Base)} & 7.38 & 9.62 & 13.81 & 25.03 & 26.55 & 7.13 & 9.30 & 13.63 & 24.86 & 26.03 \\ 
\multicolumn{1}{r|}{1} & 9.77 & 12.41 & 17.15 & 28.47 & 31.09 & 9.41 & 12.14 & 17.36 & 29.32 & 31.16 \\ 
\multicolumn{1}{r|}{2} & 10.62 & 13.13 & 18.19 & 29.42 & 32.54 & 10.50 & 13.39 & 18.76 & 30.57 & 32.09 \\ 
\multicolumn{1}{r|}{5} & \textbf{12.80} & \textbf{16.03} & \textbf{21.60} & 33.56 & \textbf{35.86} & \textbf{11.85} & \textbf{15.16} & \textbf{21.56} & \textbf{34.56} & \textbf{35.31} \\ 
\multicolumn{1}{r|}{10} & 12.14 & 15.26 & 20.78 & 32.21 & 34.77 & 11.75 & 15.01 & 21.26 & 33.90 & 34.33 \\ 
\multicolumn{1}{r|}{20} & 12.19 & 15.54 & 21.50 & \textbf{33.69} & 35.42 & 11.56 & 14.89 & 20.91 & 33.44 & 34.81 \\ \bottomrule
\end{tabular}%
}
\label{tab:gaussian_noise}
\end{table*}


\section{Quantitative Evaluation}
\label{sec:quant}

In this section, we present a thorough quantitative evaluation of our \ac{slp} model, providing results and subsequent discussion. We first conduct experiments using the \textit{Text to Gloss} setup. We then evaluate the \textit{Gloss to Pose} and the end-to-end \textit{Text to Pose} setups. Finally, we provide results of our user study with Deaf participants.

\subsection{Text to Gloss Translation}

To provide a baseline, our first experiment evaluates the performance of a classic transformer architecture \citep{vaswani2017attention} for the translation of spoken language to sign glosses sequences. We train a vanilla transformer model to predict sign gloss intermediary, with 2 layers, 8 heads and an embedding size of 256. We compare our performance against Stoll \etal \citeyearpar{stoll2020text2sign}, who use an encoder-decoder network with 4 layers of 1000 \acp{gru} as a translation architecture.

Table \ref{tab:text_to_gloss_results} shows that a transformer model achieves state-of-the-art results, significantly outperforming that of Stoll \etal \citeyearpar{stoll2020text2sign}. This supports our use of the proposed transformer architecture for sign language understanding. 


\subsection{Gloss to Pose Production} \label{sec:g2p_quant}

In our next set of experiments, we evaluate our progressive transformer on the Gloss to Pose task outlined in Section \ref{sec:eval_protols}. As a baseline, we train a progressive transformer model to translate from gloss to sign pose without augmentation.

\subsubsection{Data Augmentation}

Our base model suffers from prediction drift, with erroneous predictions accumulating over time. As transformer models are trained to predict the next time-step, they are often not robust to noise in the target input. Therefore, we experiment with multiple data augmentation techniques introduced in Section \ref{sec:data_augmentation}; namely \textit{Future Prediction}, \textit{Just Counter} and \textit{Gaussian Noise}.

\paragraph{Future Prediction} 
Our first data augmentation method is conditional future prediction, requiring the model to predict more than just the next frame in the sequence. The model is trained to produce future frames between $F_{f}$ and $F_{t}$. As can be seen in Table \ref{tab:future_prediction}, prediction of multiple future frames causes an increase in model performance, from a base level of 7.38 BLEU-4 to 11.30 BLEU-4. We believe this is because the model cannot rely on just copying the previous frame to minimise the loss, but is instead required to predict the true motion with future pose predictions.

There exists a trade-off between benefit and complexity from increasing the number of predicted frames. We find the best performance comes from a prediction of 5 frames from the current time step. This is sufficient to encourage forward planning and motion understanding, but without a large averse effect on model complexity. 

\paragraph{Just Counter}
Inspired by the memorisation capabilities of transformer models, we next evaluate a pure memorisation approach. Only the counter values are provided as target input to the model, as opposed to the usual full 3D skeleton joint positions. We show a further performance increase with this approach, considerably increasing the BLEU-4 score as shown in Table \ref{tab:just_counter}.

We believe the just counter model helps to allay the effect of drift, as the model must learn to decode the target sign pose solely from the counter position. It cannot rely on the ground truth joint embeddings it previously had access to. This halts the effect of erroneous sign pose prediction, as they are no longer fed back into the model. The setup at training and inference is now identical, requiring the model to only generalise to new data.
\paragraph{Gaussian Noise}
Our final augmentation evaluation examines the effect of applying noise to the skeleton pose sequences during training. For each joint, randomly sampled noise is applied to the input multiplied by a noise factor, $r_{n}$, representing the degree of noise augmentation.

Table \ref{tab:gaussian_noise} shows that Gaussian Noise augmentation achieves strong performance, with $r_{n} = 5$ giving the best results so far of 12.80 BLEU-4. A small amount of input noise causes the model to become more robust to auto-regressive prediction errors, as it must learn to correct the augmented inputs back to the target outputs. However, an increase of $r_{n}$ above 5 causes a large degradation, affecting the model training and subsequent testing performance.

Overall, the proposed data augmentation techniques have been shown to significantly improve model performance and are fundamental to the production of understandable sign pose sequences. In the rest of our experiments, we use Gaussian Noise augmentation with $r_{n} = 5$.

\begin{table*}[t!]
\centering
\caption{Adversarial Training results on the Gloss to Pose task. Evaluation upon inclusion of conditioning on the source input (Con.) and the amount of discriminator layers, $N$.}
\resizebox{0.9\linewidth}{!}{%
\begin{tabular}{@{}p{1.2cm}p{1.2cm}ccccc|ccccc@{}}
\toprule
 & & \multicolumn{5}{c}{DEV SET} & \multicolumn{5}{c}{TEST SET} \\ 
\multicolumn{1}{c|}{$N$} & \multicolumn{1}{c|}{Con.} & BLEU-4         & BLEU-3         & BLEU-2         & BLEU-1         & ROUGE          & BLEU-4     & BLEU-3         & BLEU-2         & BLEU-1         & ROUGE          \\ \midrule

\multicolumn{1}{c|}{6} & \multicolumn{1}{c|}{} & 12.65 & 16.09 & 22.04 & {\B 35.95} & 36.29 & 12.05 & 15.34 & 21.25 & 33.37 & 34.90 \\ \midrule

\multicolumn{1}{c|}{3} & \multicolumn{1}{c|}{\checkmark} & 12.76 & 15.91 & 21.54 & 32.97 & 36.06 & 12.16 & 15.70 & 22.34 & {\B 35.43} & 35.71 \\ 
\multicolumn{1}{c|}{4} & \multicolumn{1}{c|}{\checkmark}& 12.70 & 15.96 & 21.76 & 33.69 & 36.40 & 12.06 & 15.46 & 21.56 & 33.49 & 35.55 \\ 
\multicolumn{1}{c|}{5} & \multicolumn{1}{c|}{\checkmark} & 12.42 & 15.74 & 21.55 & 32.94 & 35.89 & 12.43 & 15.83 & 21.85 & 33.81 & 35.66 \\ 
\multicolumn{1}{c|}{6} & \multicolumn{1}{c|}{\checkmark} & {\B 13.13} & {\B 16.53} & {\B 22.36} & 34.13 & {\B 36.45} & 12.60 & 16.05 & {\B 22.37} & 34.67 & {\B 36.29} \\  
\multicolumn{1}{c|}{7} & \multicolumn{1}{c|}{\checkmark} & 12.54 & 15.96 & 21.90 & 33.62 & 36.11 & {\B 12.76} & {\B 16.15} & 22.24 & 34.36 & 35.29 \\ 
\multicolumn{1}{c|}{8} & \multicolumn{1}{c|}{\checkmark} & 12.41 & 15.89 & 22.02 & 34.99 & 35.95 & 12.38 & 15.80 & 22.09 & 34.60 & 35.85 \\ 
\bottomrule
\end{tabular}%
}
\label{tab:adv_gloss_to_pose}
\end{table*}

\begin{table*}[t!]
\centering
\caption{Mixture Density Network results on the Gloss to Pose task. Evaluation upon the mixture components, $M$ and the addition of adversarial loss (Adv.).}
\resizebox{0.9\linewidth}{!}{%
\begin{tabular}{@{}p{1.2cm}p{1.2cm}|ccccc|ccccc@{}}
\toprule
   &  & \multicolumn{5}{c}{DEV SET}  & \multicolumn{5}{c}{TEST SET} \\ 
 \multicolumn{1}{c}{$M$} & \multicolumn{1}{c|}{Adv.} & BLEU-4         & BLEU-3         & BLEU-2         & BLEU-1         & ROUGE          & BLEU-4         & BLEU-3         & BLEU-2         & BLEU-1         & ROUGE          \\ \midrule
\multicolumn{1}{c}{1} & \multicolumn{1}{c|}{} & 12.22 & 15.47 & 21.15 & 32.91 & 35.39 & 10.88 & 14.04 & 19.87 & 32.75 & 32.95 \\ 
\multicolumn{1}{c}{2} & \multicolumn{1}{c|}{}& 12.89 & 16.16 & 21.80 & 33.23 & 36.16 & 11.60 & 14.71 & 20.40 & 32.18 & 34.31 \\ 
\multicolumn{1}{c}{4} & \multicolumn{1}{c|}{}& {\B 13.14} & {\B 16.77} & {\B 22.59} & 33.84 & {\B 39.06} & 11.94 & 15.22 & 21.19 & 33.66 & 35.19 \\ 
\multicolumn{1}{c}{5} & \multicolumn{1}{c|}{}& 12.75 & 15.91 & 21.40 & 32.67 & 36.04 & 11.57 & 14.77 & 20.66 & 32.69 & 34.48 \\ 
\multicolumn{1}{c}{10} & \multicolumn{1}{c|}{}& 11.48 & 14.52 & 19.92 & 31.62 & 33.67 & 10.90 & 14.02 & 19.77 & 32.15 & 33.39 \\ 
\multicolumn{1}{c}{20} & \multicolumn{1}{c|}{}& 12.59 & 16.02 & 22.17 & {\B 35.07} & 36.28 & 12.15 & 15.35 & 21.34 & 33.62 & {\B 35.47} \\
\multicolumn{1}{c}{30} & \multicolumn{1}{c|}{}& 12.61 & 15.93 & 21.72 & 33.72 & 36.28 & 12.11 & 15.54 & 21.69 & 33.30 & 35.26 \\ 
\multicolumn{1}{c}{50} & \multicolumn{1}{c|}{}& 11.15 & 14.18 & 19.66 & 30.95 & 33.58 & 10.56 & 13.67 & 19.60 & 32.62 & 33.30 \\ \midrule
\multicolumn{1}{c}{4} & \multicolumn{1}{c|}{\checkmark}& 12.88 & 16.17 & 21.83 & 33.50 & 35.60 & {\B 12.32} & {\B 15.62} & {\B 21.82} & {\B 34.35} & 35.36 \\ \bottomrule
\end{tabular}%
}
\label{tab:mdn_g2p}
\end{table*}

\subsubsection{Adversarial Training} \label{sec:adv_quant}

We next evaluate our adversarial training regime outlined in Section~\ref{sec:adversarial}. During training, a generator, $G$, and discriminator, $D$ compete in a min-max game where $G$ must create realistic sign pose productions to fool $D$. During testing, we drop $D$ and use the trained $G$ to produce sign pose sequences given an input source text. For the adversarial experiments, we build our progressive transformer generator with 2 layers, 2 heads and an embedding size of 256. Best performance is achieved when the regression, $\lambda_{Reg}$, and adversarial, $\lambda_{GAN}$, losses are weighted as $\lambda_{Reg} = 100$ and $\lambda_{GAN} = 0.001$ respectively. This reflects the larger relative scale of the adversarial loss.

We first conduct an experiment with a non-conditional adversarial training regime. Only the sign pose sequence is critiqued, without conditioning upon source input. As shown on the top row of Table \ref{tab:adv_gloss_to_pose}, this discriminator architecture produces a weak performing generator, of only 12.65 BLEU-4. This is less than the previous augmentation results, showing how an adversary applied solely to produced sign sequences negatively affects performance. The discriminator is prompting realistic production with no regards to source text, affecting the quality of the central translation task.

We next evaluate the conditional adversarial training regime, re-introducing a critique conditioned on source input. We evaluate different discriminator architectures by varying the number of CNN layers, $N$. This changes the strength of the adversary, which is required to be finely balanced against the generator in the min-max setup. Results are shown in Table~\ref{tab:adv_gloss_to_pose}, where an increase of $N$ from 3 to 6 increases performance to a peak of 13.13 BLEU-4. This shows how a stronger discriminator can enforce a more realistic and expressive production from the generator. However, once $N$ increases further and the discriminator becomes too strong, generator performance is negatively affected. 

Overall, our conditional adversarial training regime has demonstrated improved performance over a model trained solely with a regression loss. Even for the test set, the result of 12.76 BLEU-4 is considerably higher than previous performance. This shows that the inclusion of a discriminator model increases the comprehension of sign production when conditioned on source sequence input. We believe this is due to the discriminator pushing the generator towards both a more expressive production and an accurate translation, in order to deceive the adversary. This, in turn, increases the sign content contained in the generated sequence, leading to a more understandable output and higher performance.




\begin{table*}[t!]
\centering
\caption{Results of the Text to Pose task for different model configurations.}
\resizebox{0.9\linewidth}{!}{%
\begin{tabular}{@{}p{2.8cm}ccccc|ccccc@{}}
\toprule
 & \multicolumn{5}{c}{DEV SET} & \multicolumn{5}{c}{TEST SET} \\ 
\multicolumn{1}{r|}{Configuration:} & BLEU-4         & BLEU-3         & BLEU-2         & BLEU-1         & ROUGE          & BLEU-4         & BLEU-3         & BLEU-2         & BLEU-1         & ROUGE          \\ \midrule
\multicolumn{1}{r|}{Base} & 7.30 & 9.21 & 12.87 & 23.15 & 26.11 & 6.79 & 8.74 & 12.57 & 23.46 & 25.02 \\

\multicolumn{1}{r|}{Gaussian Noise} & 10.75 & 13.47 & 18.41 & 29.43 & 32.02 & 10.08 & 12.91 & 18.17 & 29.96 & 31.66 \\

\multicolumn{1}{r|}{Adversarial} & 11.41 & 14.26 & 19.45 & {\B 31.02} & {\B 33.59} & 10.16 & 12.98 & 18.33 & 29.61 & 32.03 \\
\multicolumn{1}{r|}{\ac{mdn}} & {\B 11.54} & {\B 14.48} & {\B 19.63} & 30.94 & 33.40 & {\B 11.68} & {\B 14.55} & {\B 19.70} & {\B 31.56} & 33.19 \\
\multicolumn{1}{r|}{\ac{mdn} + Adv.} & 11.49 & 14.36 & 19.38 & 30.04 & 33.92 & 11.18 & 14.08 & 19.35 & 30.66 & {\B 33.43}  \\
\bottomrule
\end{tabular}%
}
\label{tab:t2p_configs}
\end{table*}

\begin{table*}[t!]
\centering
\caption{Results of the \textit{Text to Pose} and \textit{Text to Gloss to Pose} network configurations for the Text to Pose task.}
\resizebox{0.9\linewidth}{!}{%
\begin{tabular}{@{}p{2.8cm}ccccc|ccccc@{}}
\toprule
 & \multicolumn{5}{c}{DEV SET} & \multicolumn{5}{c}{TEST SET} \\ 
\multicolumn{1}{c|}{Configuration:} & BLEU-4         & BLEU-3         & BLEU-2         & BLEU-1         & ROUGE          & BLEU-4         & BLEU-3         & BLEU-2         & BLEU-1         & ROUGE          \\ \midrule
\multicolumn{1}{r|}{Text to Pose}  &  {\B 11.54} & {\B 14.48} & {\B 19.63} & {\B 30.94} & {\B 33.40} &  11.68 & 14.55 & 19.70 & 31.56 & 33.19 \\
\multicolumn{1}{r|}{Text to Gloss to Pose} & 11.21 & 14.22 & 19.46 & 30.37 & 32.95 & {\B 13.64} & {\B 17.05} & {\B 23.09} & {\B 34.94} & {\B 36.90} \\ \bottomrule
\end{tabular}%
}
\label{tab:configuration_results}
\end{table*}

\subsubsection{Mixture Density Networks}

Our final Gloss to Pose evaluation is of the \acf{mdn} model configuration outlined in Section \ref{sec:mdn}. During training, a multimodal distribution is created that best models the data, which is then used to sample from during inference. In this experiment, our progressive transformer model is built with 2 layers, 2 heads and an embedding size of 512.

We evaluate different numbers of mixture components, $M$, with results shown in Table \ref{tab:mdn_g2p}. As shown, initially increasing $M$ allows a multimodal prediction over a larger subspace, better modelling the sequence variation. This is supported by the results, with $M = 4$ achieving the highest validation performance of 13.14 BLEU-4. We find the regression to the mean of a deterministic prediction to be reduced, leading to a more expressive production. The subtleties of sign poses are restored, particularly for the small and variable finger joints. As $M$ increases further, the added model complexity outweighs these benefits, leading to a performance degradation. 

Our proposed \ac{mdn} formulation achieves a higher performance than the previous deterministic approach of the progressive transformer. Comparison against the adversarial configuration shows a slight increase in performance (13.14 and 13.13 BLEU-4 respectively).  However, given the back translation evaluation is not perfect, one might consider the performance of the \ac{mdn} and adversarial models' to be similar, within the error margin of the \ac{slt} system. Both methods have a similar result of reducing the regression to the mean found in the original architecture and increasing sign pose articulation.

We additionally evaluate the combination of the \ac{mdn} loss with the previously described adversarial loss, as explained in Section \ref{sec:mdn_adv}. This creates a network that uses a mixture distribution generator and a conditional discriminator. As in Section \ref{sec:adv_quant}, we weight the \ac{mdn}, $\lambda_{MDN} = 100$, and adversarial, $\lambda_{GAN} = 0.001$, losses respectively. As shown at the bottom of Table \ref{tab:mdn_g2p}, a combination of the \ac{mdn} and adversarial training actually results in a lower performance than either individually on the dev set, of 12.88 BLEU-4. However, for the test set, this combination results in a slightly better performance than the \ac{mdn} alone. Both of these configurations aim to alleviate the effect of regression to the mean, but may adversely affect the performance of the other due to their similar goals.

\subsection{Text to Pose Production}

We next evaluate our models on the Text to Pose task outlined in Section \ref{sec:eval_protols}. This is the true end-to-end translation task, direct from a source spoken language sequence without the need for a gloss intermediary.

\subsubsection{Model Configurations}

We start by evaluating the various model configurations proposed in Section \ref{sec:methodology}; namely base architecture, Gaussian noise augmentation, adversarial training and the \ac{mdn}. The results of different configurations are shown in Table \ref{tab:t2p_configs}.

As with the Gloss to Pose task, Gaussian Noise augmentation increases performance from the base architecture, from 7.30 BLEU-4 to 10.75. We believe this is due to the reduction of the prediction drift as previously explained. The addition of adversarial training again increases performance, to 11.41 BLEU-4. The conditioning of the discriminator is even more important for this task, as the input is spoken language and provides more context for production.

The best Text to Pose performance of 11.54 BLEU-4 comes from the \ac{mdn} model. 
As mentioned earlier, the performance of the adversarial and \ac{mdn} setups' can be seen as equivalent considering the utilized \ac{slt} system is not perfect. Due to the increased context given by the source spoken language, there is a larger natural variety in sign production. Therefore, the multimodal modelling of the \ac{mdn} is further enhanced, as highlighted by the performance gains. The addition of adversarial training on top of an \ac{mdn} model does not increase performance further, as was seen in the previous evaluations.

\subsubsection{Text to Pose v Text to Gloss to Pose}

Our final experiment evaluates two end-to-end network configurations; sign production either direct from text (Text to Pose (T2P)) or via a gloss intermediary (Text to Gloss to Pose (T2G2P)). These two tasks are outlined in Figure \ref{fig:SLP_Text_Gloss_Seq}, T2G2P on the left, T2P on the right.

As can be seen from Table \ref{tab:configuration_results}, the T2P model outperforms the T2G2P for the development set. We believe this is because there is more information available within spoken language compared to a gloss representation, with more tokens per sequence to predict from. Predicting gloss sequences as an intermediary can act as an information bottleneck, as all the information required for production needs to be present in the gloss. Therefore, any contextual information present in the source text can be lost. However, in the test set, we achieve better performance using gloss intermediaries. We believe this is due to the effects of the limited number of training samples and the smaller vocabulary size of glosses on the generalisation capabilities of our networks.

The success of the T2P network shows that our progressive transformer model is powerful enough to complete two sub-tasks; firstly mapping spoken language sequences to a sign representation, then producing an accurate sign pose recreation. This is important for future scaling of the \ac{slp} model architecture, as many sign language domains do not have gloss availability. 

Furthermore, our final BLEU-4 scores outperform similar end-to-end Sign to Text methods which do not utilise gloss information \citep{camgoz2018neural} (9.94 BLEU-4). Note that this is an unfair direct comparison, but it does provide an indication of model performance and the quality of the produced sign pose sequences.

\subsection{User Evaluation}

The only true way to evaluate the sign production is in discussion with the Deaf communities, the end users. As our outputs are sign language sequences, we wish to understand how understandable they are to a native Deaf signer. We perform this evaluation with the skeletal output of the model, as we do not wish to confuse the translation ability of the system with the visual aesthetics of an avatar. However, by assessing the skeleton directly, we lose a lot of information that is conveyed in images such as shadow and occlusion. We therefore do a relative comparison between ground-truth and produced sequences, allowing us to assess the productions fairly. Although this work is in its infancy, we understand it is important to get early feedback from the Deaf communities. We believe the Deaf communities should be empowered and be involved in all steps of the development of any technology that is targeting their native languages. 

We conducted a user evaluation with native \ac{dgs} speakers to estimate the comprehension of our produced sign pose sequences. We designed a survey consisting of a comparison of the productions against ground truth data, the \textit{Visual Task}, and a \textit{Translation Task} that evaluates the sign comprehension. We animated our sign pose sequences as explained in Section \ref{sec:sign_seq_outputs} and placed the videos in an online survey. The user evaluation was conducted in collaboration with \textit{HFC Human-Factors-Consult GmbH}.

We evaluated with two different model configurations; adversarial training and \acp{mdn}, providing users with different sequences from each and randomising the order of the videos. We received 20 Deaf participants who completed the evaluation, both comparing the production quality and testing the sign comprehension. 
\begin{figure*}[h]
    \centering
    \includegraphics[width=0.9 \linewidth]{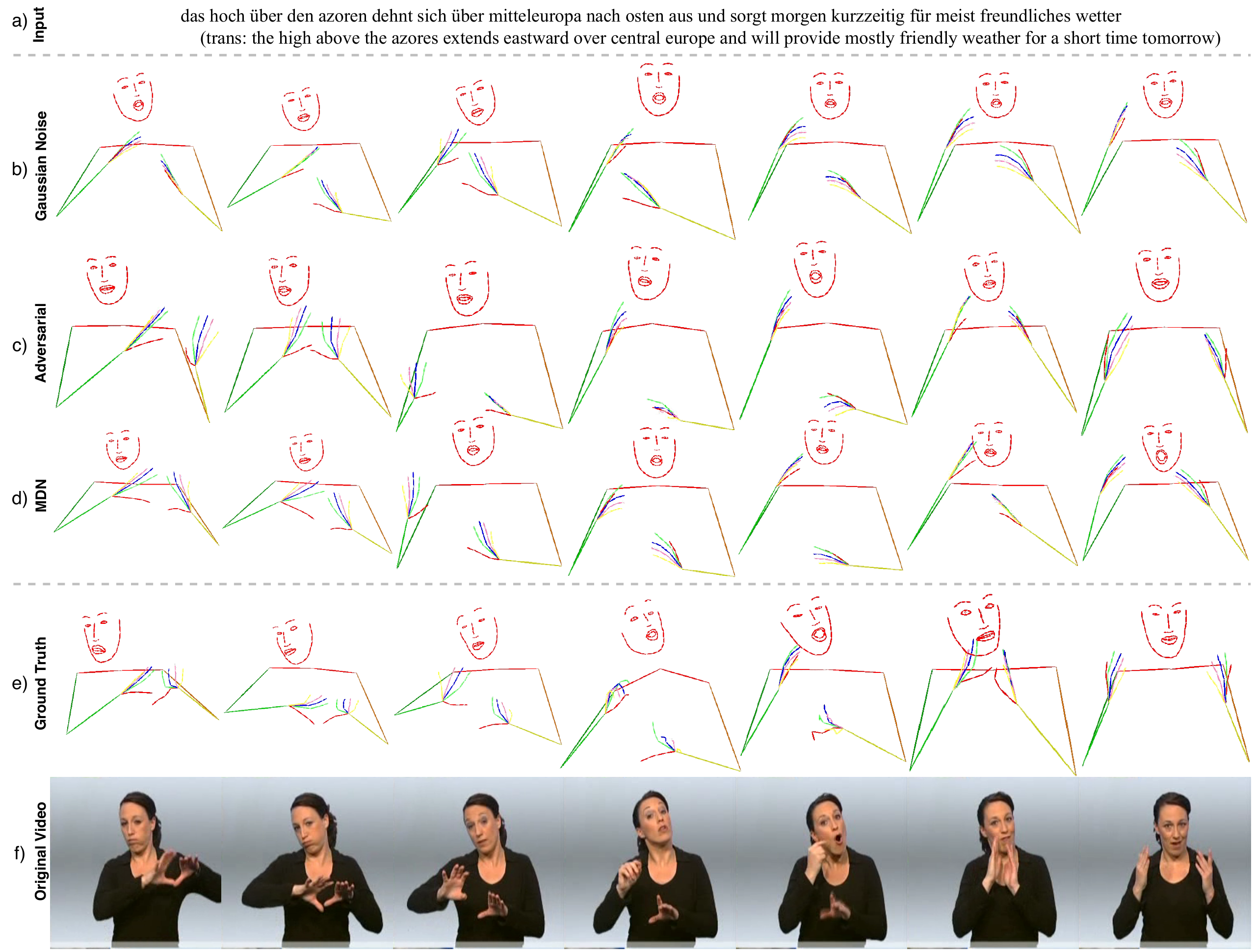}
    \caption{Qualitative evaluation of an example sign pose sequence. The source input is at the top, with the ground truth video frames and poses at the bottom. Middle rows contain produced sign pose sequences of different model configurations.}
    \label{fig:qual_eval_7496}
\end{figure*}%

\subsubsection{Visual Task}

Our first evaluation is a visual task, where a video of a sign production is shown alongside the corresponding ground truth sign sequence. The user is asked to rate both videos, with an implicit comparison between them. The comparison results are shown in Table \ref{tab:user_eval_visual}, for both the adversarial and \ac{mdn} model configurations.

\begin{table}[h]
\centering
\caption{User evaluation results of the Visual task, showing the percentage of users who rated the ground truth (GT) or produced sequences (Prod) of a higher visual quality or equal..}
\resizebox{0.8\linewidth}{!}{%
\begin{tabular}{@{}p{2.0cm}|ccc@{}}
\toprule
Configuration: & GT & Prod & Equal \\ \midrule
Adversarial  & 14.58\% & 8.33\% & 77.08\% \\
\ac{mdn}  & 0.00\% & 15.38\% & 84.62\% \\
\bottomrule
\end{tabular}%
}
\label{tab:user_eval_visual}
\end{table}

Overall, the user feedback was mainly equal between the produced and ground-truth videos, with slightly more participants preferring the productions. This highlights the quality of the produced sign language videos, often as they are smoothly generated without any visual jitters. On the contrary, the original sequences often suffer from visual jitter, due to the motion blur in the original videos and the artifacts introduced in the 3D pose estimation.

The \ac{mdn} configuration received higher ratings from the participants than the adversarial setup. 15.38\% of users preferred the \ac{mdn} productions over the ground-truth sequences, compared to 8.33\% for the adversarial model. This demonstrates that the participants preferred the visuals of the \ac{mdn} model. The quantitative back translation results for these models were similar (Section \ref{sec:g2p_quant}), but the users feedback suggests the \acp{mdn} production was of higher quality. 

\begin{figure}[b]
    \centering
    \includegraphics[width=0.85 \linewidth]{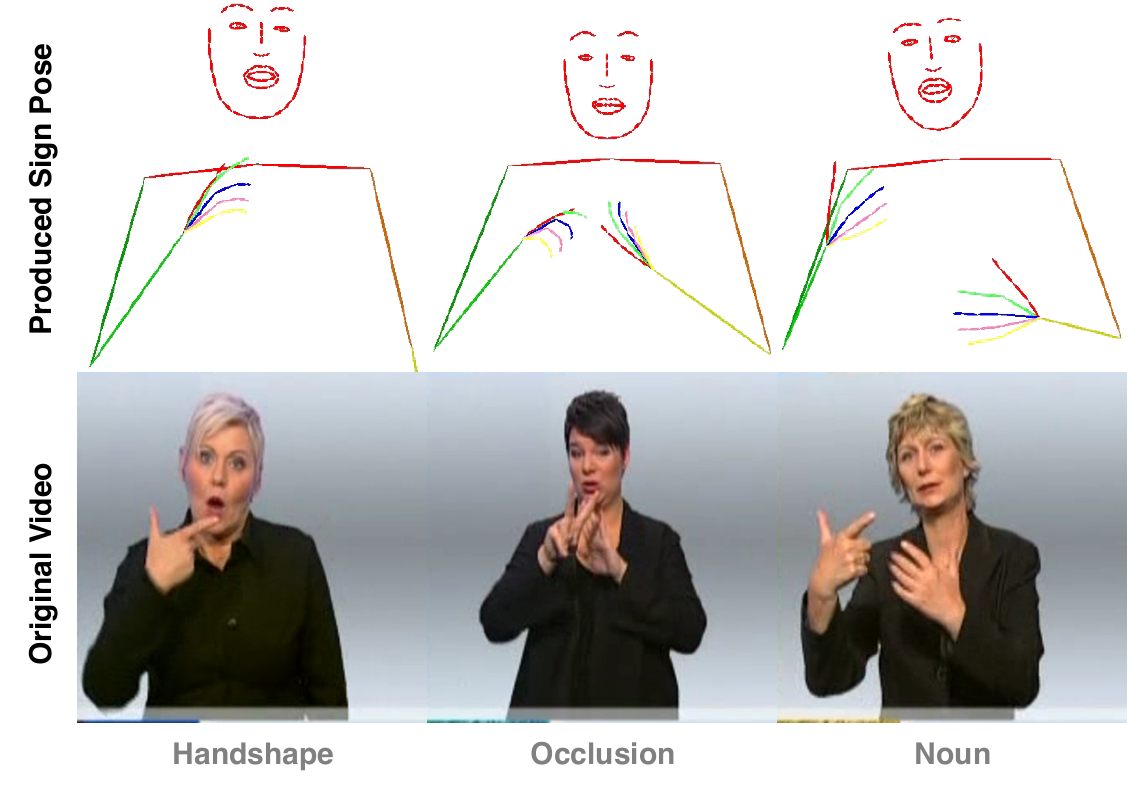}
    \caption{Example failure sign pose productions. Due to either complex handshape (left), hand occlusion (middle) or proper noun (right).}
    \label{fig:qual_failures}
\end{figure}%

\subsubsection{Translation Task}

Our second evaluation is a translation task, designed to measure the translation accuracy of the sign productions. An automatic production was shown alongside 4 possible spoken language translations of the sign sequence, where one is the correct sentence. The user is asked to select the most likely translation.

\begin{table}[h]
\centering
\caption{User evaluation results for the Translation task, showing the percentage of participants who chose the correct spoken language translation out of a choice of 4.}
\resizebox{0.5\linewidth}{!}{%
\begin{tabular}{@{}p{2.0cm}|ccc@{}}
\toprule
Configuration: & Correct \\ \midrule
Adversarial  & 34.72\% \\
\ac{mdn}  & 78.57\%  \\
\bottomrule
\end{tabular}%
}
\label{tab:user_eval_transaltion}
\end{table}

Table \ref{tab:user_eval_transaltion} shows that, for the adversarial examples, 34.72\% of users chose the correct translation, compared to 78.57\% for the \ac{mdn} configuration. This is a drastic difference in the understanding of each of the model configurations, further demonstrating the success of the \ac{mdn} productions. With the results of both visual and translation tasks, alongside the similar quantitative performance, we can conclude that the proposed \ac{mdn} configuration generates the most realistic and expressive sign pose production.

\section{Qualitative Evaluation}
\label{sec:qual}

\begin{figure*}[b]
    \centering
    \includegraphics[width=0.9 \linewidth]{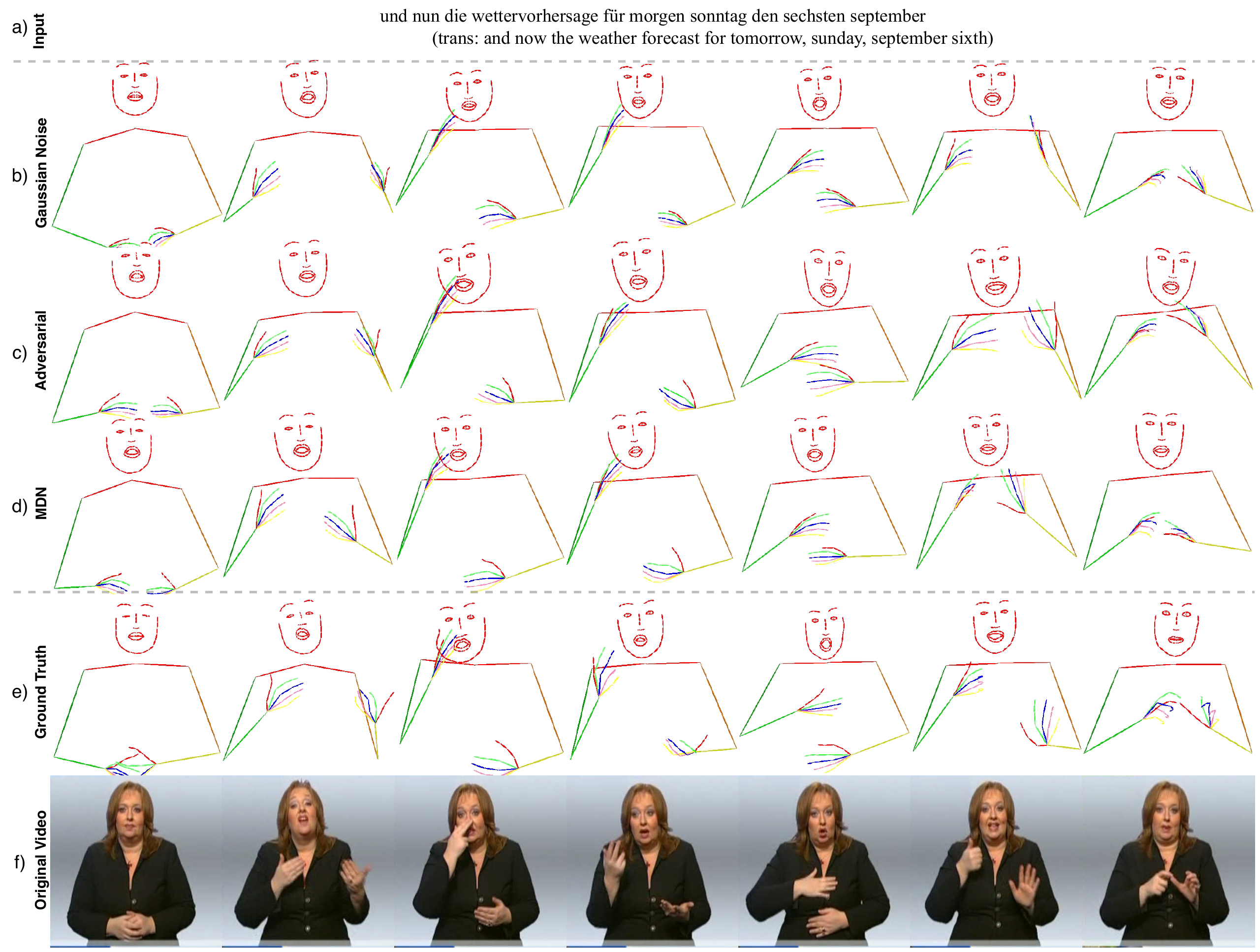}
    \caption{Qualitative evaluation of an example sign pose sequence. The source input is at the top, with the ground truth video frames and poses at the bottom. Middle rows contain produced sign pose sequences of different model configurations.}
    \label{fig:qual_eval_4078}
\end{figure*}%

In this section, we report qualitative results for our \ac{slp} model. We share snapshot examples of sign pose sequences in Figures \ref{fig:qual_eval_7496} and \ref{fig:qual_eval_4078}, visually comparing the outputs of the proposed model configurations for the gloss to pose task. The corresponding unseen spoken language sequence is shown as input at the top, alongside example frames from the ground truth video and the produced sign language sequence.

As can be seen from the provided examples, our \ac{slp} model produces visually pleasing and realistic looking sign with a close correspondence to the ground truth video. Body motion is smooth and accurate, whilst hand shapes are meaningful if a little under-expressed. Specific to non-manual features, we find a close correspondence to the ground truth video alongside accurate head movement, with a slight under-articulation of mouthings.

For comparisons between model configurations, the Gaussian Noise productions can be seen to be under-expressed, specifically the hand shape and motions of Figure \ref{fig:qual_eval_7496}b. The adversarial training improves this, resulting in a significantly more expressive production, with larger hand shapes seen in the 6th frame of Figure \ref{fig:qual_eval_4078}c. This is due to the discriminator pushing the productions towards a more realistic output. Inclusion of a \ac{mdn} representation can be seen to provide more accuracy in production, with the sign poses of Figure \ref{fig:qual_eval_7496}d visually closer to the ground truth. This is due to the mixture distribution modelling the uncertainty of the continuous sign sequences, removing the mean productions that can be seen in the Gaussian Noise productions.

Visual comparisons between the adversarial and \ac{mdn} productions reflect the equal quantitative performance of the two (Section \ref{sec:g2p_quant}), demonstrating two contrasting ways of increasing the sign comprehension. Overall, the problem of regression to the mean is diminished and a more realistic production is achieved, highlighting the importance of the proposed model configurations.

These examples show that regressing continuous 3D human pose sequences can be successfully achieved using a self-attention based approach. The predicted joint locations for neighbouring frames are closely positioned, showing that the model has learnt the subtle signer movements. Smooth transitions between signs are produced, highlighting a difference from the discrete generation of spoken language. 

Figure \ref{fig:qual_failures} shows some failure cases of the approach. Complex hand classifiers can be difficult to replicate (left) and hand occlusion affects the quality of training data (middle). We find that the most difficult production occurs with proper nouns and specific entities, due to the lack of grammatical context and examples in the training data (right).

\section{Conclusions}
\label{sec:conc}

In this work, we presented a Continuous 3D Multi-Channel Sign Language Production model, the first \ac{slp} model to translate from text to continuous 3D sign pose sequences in an end-to-end manner. To enable this, we proposed a \textit{Progressive Transformer} architecture with an alternative formulation of transformer decoding for variable length continuous sequences. We introduced a counter decoding technique to predict continuous sequences of variable lengths by tracking the production progress over time and predicting the end of sequence. 

To reduce the prediction drift that is often seen in continuous sequence production, we presented several data augmentation methods that significantly improve model performance. Predicting continuous values often results in under-articulated output, and thus we proposed the addition of adversarial training to the network, introducing a conditional discriminator model to prompt a more realistic and expressive production. We also proposed a \acf{mdn} modelling, utilising the progressive transformer outputs to paramatise a mixture Gaussian distribution.

We evaluated our approach on the challenging \ac{ph14t} dataset, proposing a back translation evaluation metric for \ac{slp}. Our experiments showed the importance of data augmentation techniques to reduce model drift. We improved our model performance with the addition of both an adversarial training regime and a \ac{mdn} output representation. Furthermore, we have shown that a direct text to pose translation configuration can outperform a gloss intermediary model, meaning \ac{slp} models are not limited to domains where expensive gloss annotation is available. 

Finally, we conducted a user study of the Deaf's response to our sign productions, understanding the sign comprehension of the proposed model configurations. The results show that our productions, while not perfect, can be further improved by reducing and smoothing noise inherent to the data and approaches. However, they also highlight that the current sign productions still need improvement to be fully understandable by the Deaf. The field of \ac{slp} is in its infancy, with a potential for large growth and improvement in the future. 

We believe the current 3D skeleton representation affects the comprehension of sign pose sequences. As future work, we would like to increase the realism of sign production by generating photo-realistic signers, using \ac{gan} image-to-image translation models \citep{isola2017image,zhu2017unpaired,chan2019everybody} to expand from the current skeleton representation. Drawing on feedback from the user evaluation, we plan to improve the hand articulation via a hand shape classifier to increase comprehension. An automatic viseme generator could also be included to the pipeline to improve mouthing patterns, producing features in a deterministic manner direct from dictionary data.

\section{Acknowledgements}

We would like to thank Tao Jiang for their help with data curation. This work received funding from the SNSF Sinergia project ‘SMILE’ (CRSII2 160811), the European Union’s Horizon2020 research and innovation programme under grant agreement no. 762021 ‘Content4All’ and the EPSRC project ‘ExTOL’ (EP/R03298X/1). This work reflects only the authors view and the Commission is not responsible for any use that may be made of the information it contains. 






\printbibliography

\end{document}

%% file: acronyms.tex
\begin{acronym}[PHOENIX14\textbf{T} ]

\acrodefplural{rnn}[RNNs]{Recurrent Neural Networks}
\acrodefplural{cnn}[CNNs]{Convolutional Neural Networks}
\acrodefplural{hmm}[HMMs]{Hidden Markov Models}
\acrodefplural{gru}[GRUs]{Gated Recurrent Units}
\acrodefplural{crf}[CRFs]{Conditional Random Fields}
\acrodefplural{gan}[GANs]{Generative Adversarial Networks}
\acrodefplural{gpu}[GPUs]{Graphic Processing Units}

\acrodefplural{mdn}[MDNs]{Mixture Density Networks}

\acro{bsl}[BSL]{British Sign Language}
\acro{bleu}[BLEU]{Bilingual Evaluation Understudy}
\acro{blstm}[BLSTM]{Bidirectional Long Short-Term Memory}
\acro{cnn}[CNN]{Convolutional Neural Network}
\acro{crf}[CRF]{Conditional Random Field}
\acro{cslr}[CSLR]{Continuous Sign Language Recognition}
\acro{ctc}[CTC]{Connectionist Temporal Classification}
\acro{dl}[DL]{Deep Learning}
\acro{dgs}[DGS]{German Sign Language - Deutsche Gebärdensprache}
\acro{dsgs}[DSGS]{Swiss German Sign Language - Deutschschweizer Geb\"ardensprache}
\acro{dtw}[DTW]{Dynamic Time Warping}
\acro{fc}[FC]{Fully Connected}
\acro{ff}[FF]{Feed Forward}
\acro{gan}[GAN]{Generative Adversarial Network}
\acro{gpu}[GPU]{Graphics Processing Unit}
\acro{gru}[GRU]{Gated Recurrent Unit}
\acro{gtpt}[G2PT]{Gloss to Pose Transformer}
\acro{hmm}[HMM]{Hidden Markov Model}
\acro{isl}[ISL]{Irish Sign Language}
\acro{lstm}[LSTM]{Long Short-Term Memory}
\acro{mha}[MHA]{Multi-Headed Attention}
\acro{mtc}[MTC]{Monocular Total Capture}
\acro{mse}[MSE]{Mean Squared Error}
\acro{mdn}[MDN]{Mixture Density Network}
\acro{nmt}[NMT]{Neural Machine Translation}
\acro{nlp}[NLP]{Natural Language Processing}
\acro{ph12}[PHOENIX12]{RWTH-PHOENIX-Weather-2012}
\acro{ph14}[PHOENIX14]{RWTH-PHOENIX-Weather-2014}
\acro{ph14t}[PHOENIX14\textbf{T}]{RWTH-PHOENIX-Weather-2014\textbf{T}}
\acro{pof}[POF]{Part Orientation Field}
\acro{paf}[PAF]{Part Affinity Field}
\acro{pttt}[P2TT]{Pose to Text Transformer}
\acro{relu}[RELU]{Rectified Linear Units}
\acro{rnn}[RNN]{Recurrent Neural Network}
\acro{rouge}[ROUGE]{Recall-Oriented Understudy for Gisting Evaluation}
\acro{sgd}[SGD]{Stochastic Gradient Descent}
\acro{sla}[SLA]{Sign Language Assessment}
\acro{slr}[SLR]{Sign Language Recognition}
\acro{slt}[SLT]{Sign Language Translation}
\acro{slp}[SLP]{Sign Language Production}
\acro{smt}[SMT]{Statistical Machine Translation}

\acro{ttgt}[T2GT]{Text to Gloss Transformer}
\acro{ttpt}[T2PT]{Text to Pose Transformer}
\acro{ttp}[T2P]{Text to Pose}
\acro{ttgtp}[T2G2P]{Text to Gloss to Pose}
\acro{wer}[WER]{Word Error Rate}

\end{acronym}